\documentclass[jair, 11pt]{article}
\usepackage{theapa}
\usepackage{jair}
\usepackage{amsmath}
\usepackage{amssymb}
\usepackage{amsthm}
\usepackage{newtxmath}
\usepackage[T1]{fontenc}
\usepackage{setspace}
\usepackage{mathtools}
\usepackage[implicit=false]{hyperref}
\usepackage{url}
\usepackage{booktabs}
\usepackage{nicefrac}
\usepackage{microtype}
\usepackage{graphicx}
\usepackage{subcaption}
\usepackage{accents}
\usepackage{physics}
\usepackage{xfrac}
\usepackage{pgf}
\usepackage{relsize}

\newtheorem{theorem}{Theorem}
\newtheorem{lemma}[theorem]{Lemma} % same counter as theorem
\newtheorem{proposition}{Proposition}
\theoremstyle{definition}
\theoremstyle{remark}

\jairheading{73}{2022}{1323--1353}{01/2022}{04/2022}
\ShortHeadings{Point Process Metric Space}
{Marmarelis, Ver Steeg, \& Galstyan}
\firstpageno{1323}

\begin{document}

\title{A Metric Space for Point Process Excitations}% to Recover a Contagion Process}%Effective Spatiotemporal Contagion Process}
\author{\name Myrl G. Marmarelis \email myrlm@isi.edu\\
        \name Greg Ver Steeg \email gregv@isi.edu\\
        \name Aram Galstyan \email galstyan@isi.edu\\
        \addr USC Information Sciences Institute}

\maketitle

\begin{abstract}
  A multivariate Hawkes process enables self- and cross-excitations through a triggering matrix that behaves like an asymmetrical covariance structure, characterizing pairwise interactions between the event types. Full-rank estimation of all interactions is often infeasible in empirical settings.
  Models that specialize on a spatiotemporal application alleviate this obstacle by exploiting spatial locality, allowing the dyadic relationships between events to depend only on separation in time and relative distances in real Euclidean space. Here we generalize this framework to any multivariate Hawkes process, and harness it as a vessel for embedding arbitrary event types in a \emph{hidden} metric space. Specifically, we propose a Hidden Hawkes Geometry (\texttt{HHG}) model to uncover the hidden geometry between event excitations in a multivariate point process. The low dimensionality of the embedding  regularizes the structure of the inferred interactions. We develop a number of estimators and validate the model by conducting several experiments. In particular, we investigate regional infectivity dynamics of COVID-19 in an early South Korean record and recent Los Angeles confirmed cases. By additionally performing synthetic experiments on short records as well as explorations into options markets and the Ebola epidemic, we demonstrate that learning the embedding alongside a point process uncovers salient interactions in a broad range of applications. % is resilient to ambiguities in the relationships among event types
\end{abstract}

\section{Introduction}% \& Motivation}%\footnote{observe the unraveling causal structure in Figure~\ref{fig:illustration}.}
\emph{Infectious diseases}~\cite{ref:gibson}, \emph{news topics}~\cite{ref:he}, \emph{crime patterns}~\cite{ref:mohler2011}, \emph{neuronal spike trains}~\cite{ref:pillow}, and \emph{market trade-level activity}~\cite{ref:pakkanen,ref:swishchuk} naturally suit the form of a diachronic point process with a network of directed excitations. Understanding their intrinsic dynamics is of immense scientific and strategic value: a particular series of discrete options trades may inform an observer on the fluctuating dispositions of market agents; similarly, temporal news publication patterns may betray an ensuing shift in the public zeitgeist. The spread of a novel pathogen, notably the COVID-19 virus, through disjointed pockets of the globe hints to how it proliferates, and how that might be averted~\cite{ref:drakopoulos}.

\begin{figure}[ht]\centering
 \hfill\vspace{1em}\\
 \includegraphics[width=.9\textwidth]{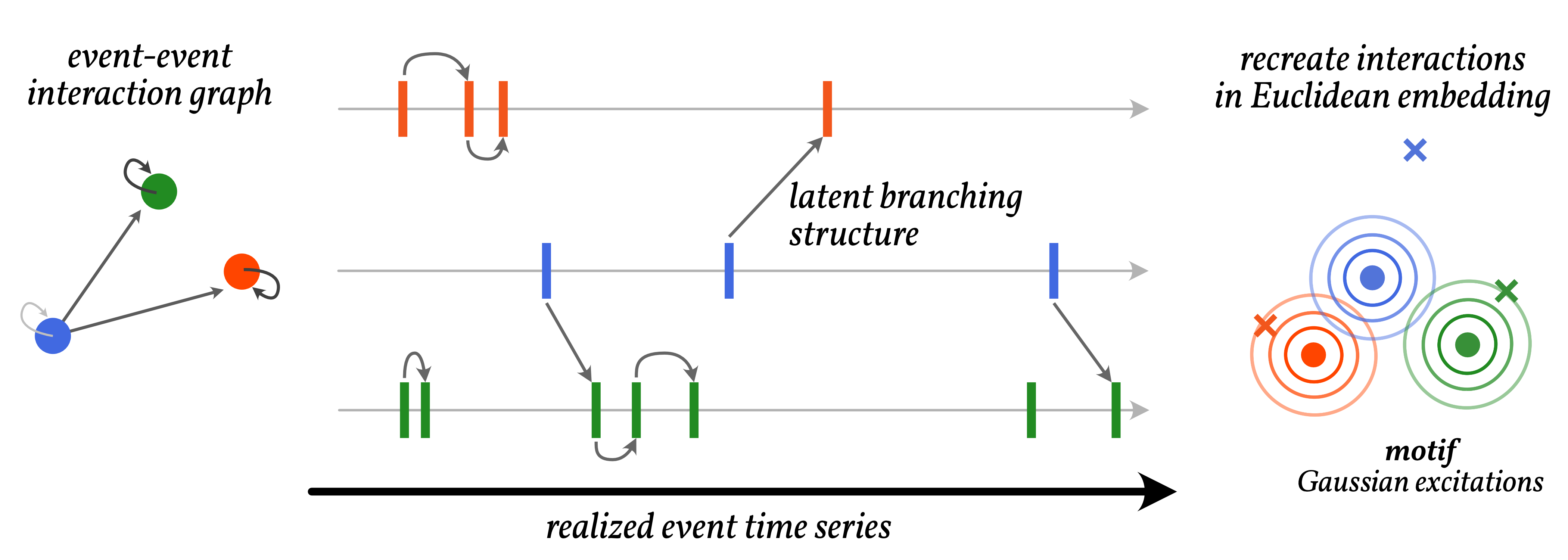}
 %\caption{\label{fig:illustration} Recovering the spatiotemporal configuration of receptions (exes) and influences (triangles) in a point process by proxy of the latent causality structure. Instead of learning all possible interactions, we estimate a Euclidean embedding with fewer degrees of freedom and the additional benefit of interpretability. Distances between influence and reception points are mostly preserved, with the influencers compressed together as an artifact of empirical confounders.}\vspace{-.5em}
 \caption{\label{fig:illustration} A stylized rendition. Instead of learning all possible interactions, we estimate a compact Euclidean embedding with fewer degrees of freedom and the additional benefit of interpretability.}
\end{figure}

%Deep nonparametric (Poisson or even intensity-free) point process models made massive strides in the past few years. However, not all domains permit such expressive characterizations. In simpler Hawkes processes,

Simple Hawkes~\citeyear{ref:hawkes} processes model excitations as a linear combination of responses to past events. Often in the multivariate setting, a drastic need for data-efficient techniques arises: practically, estimating the $(n\times n)$ possible excitations between each dyad (pair) of event types is often untenable without succinct and interpretable parametrization. How can one possibly disentangle the contributions of hundreds of options trades within each minute, in a myriad of different strike prices and expiration dates, to the future frequency of a particular type of trade? Temporal processes like markets are also highly non-stationary, exacerbating the restriction to learn meaningful models on short windows. In these cases it is necessary to envision a reduced parametric form that is expressive yet cogent.
The \emph{manifold hypothesis}~\cite[for possible references]{ref:fefferman} supposes that the vast majority of complex, high-dimensional phenomena reside in a state space of significantly lower dimensionality.
The present endeavor is an effort to estimate embeddings of manifold-like structures from indirect measurements---in this case, marked events.

Spatiotemporal domains~\cite{ref:schoenberg,ref:mohler,ref:yuan} exploit physical constraints on interaction locality. In essence, the direct influence one event bears on another is restricted by their separation in space, invariant to absolute positioning. We show that it is feasible to conjure a \emph{latent} metric representation of the event types by harnessing a spatiotemporal approach outside the spatial context. We tailor the latent space toward the precise interactions between events.% under question. %We improve a number of predictive scores with this structural imposition in applications that are both spatial and explicitly not.

Our first contribution is a model that captures multivariate point-process regularities in a compressed metric space, and which eliminates the need to estimate all the pairwise interactions between the event types. As our second contribution, we propose three estimation algorithms founded on the expectation-maximization (EM) technique in identifying Hawkes processes~\cite{ref:yuan,ref:zhou2,ref:halpin}. A latent structure introduced in the E-step to attribute the source of each event realization (Figure~\ref{fig:illustration}) permits tractable solutions in the M-step, which we exploit to optimize the embedding directly. Third, our findings on diverse social datasets demonstrate improvements upon the State of the Art (SotA) models, while also yielding qualitatively intuitive representations~\cite{ref:salehi,ref:xu,ref:zhou1}. We successfully learn the regional diffusion process for the 2014--2015 Ebola outbreak as well as the influences between options trades in different stocks. We also optimized an embedding for the South Korean early COVID-19 outbreak~\cite{ref:korea} and the winter stage of the pandemic in Los Angeles~\cite{ref:latimes}.%, showcased in Appendix C for lack of sufficient space. % preliminary? "sufficient room"?

One active line of work, paralleling ours, embeds each event based on some conceived heuristic like temporal proximity~\cite{ref:torricelli,ref:zhu2,ref:zuo}; another direction of inquiry entails the estimation of low-rank multivariate Hawkes processes~\cite{ref:nickel,ref:lemonnier,ref:junuthula}. We differ from both in learning an actual metric-space representation vis-\`a-vis the real Hawkes likelihood. To our knowledge, we are the first to propose a Euclidean embedding scheme driven entirely by an underlying marked Hawkes process.

\section{Method}
Consider a record of $N$ event occurrences $(k_i, t_i)\in\mathbb{H}$, $i=1,2,\dots N$, with $n$ marked types $k_i\in\{1,2,\dots n\}$ at times $t_i\in[0,T)$. We assume that events with certain marks excite future events of either the same or another type. Multiple such interactions may be present, and we desire to identify those that are warranted by the available observations. A compact representation would, in effect, induce a shrinkage prior on the continuum of allowable interactions. Its form should align with our inductive bias. % BE CAREFUL NOW: "LATENT" SHOULD BE THE UNOBSERVED VARIABLES, NOT PARAMETERS, I.E. THE CAUSALITIES IN OUR CONTEXT

\paragraph{We begin with preliminaries.} The multivariate intensity function $\lambda(k_j,t_j)$, conditional on the events in $[0,t_j)$, dictates the instantaneous frequency of event $(k_j,t_j)$. Any interval $[t,t+\mathrm{d}t)$ witnesses Poisson-distributed instances of event type $k$ with rate $\lambda(k,t)\mathrm{d}t$. We decompose this intensity~\cite{ref:bacry} into self- and cross-excitations and an intrinsic background rate, not knowing a priori which event triggered which: %on which specific event or type thereof triggered any particular future event. %a convolution on the spike trains by a causal kernel bank $\tau\mapsto h(k,l,\tau)$ for the response from type $l$ to type $k$. sans!
\begin{equation}\label{eq:intensity}
\lambda(k_j,t_j)\coloneqq\sum_{i=1}^N h(k_j, {k_i}, t_j - t_i) + \mu(k_j).%\sum_{(k_i,t_i)}
\end{equation}

A variety of approaches exists to estimate or infer the $(n\times n)$ separate response functions, $\tau \mapsto h(k_j, {k_i}, \tau),$ indexed by the influenc\emph{ing} event type $k_i$ and influenc\emph{ed} type $k_j$. Short records and ambiguities in the combinatorial structure of potential causation hinder proper identification of actual interactions. % real, true ?
In the presence of competing renditions for the $\mathcal{O}(n^2)$ coefficients, we elect the structure that follows from the manifold hypothesis. Concretely, we envision a Euclidean geometry that adequately captures the interactions between event types. We estimate distinct embeddings for receiving, $x_{k_j}\in X\subset\mathbb{R}^m$, versus influencing, $y_{k_i}\in Y\subset\mathbb{R}^m$, event types because otherwise we inadvertently constrain ourselves to symmetrical interactions. Multivariate Hawkes processes generally admit directed interactions. Accordingly, the finite collections of vectors $X$ and $Y$ constitute the embedding of a putative manifold wherein the multivariate component of the excitations is characterized by $\lVert y_{k_i}-x_{k_j}\rVert_2$\ ---the influence that some event type $k_i$ exerts on $k_j$. From hereon, we shall express $y_{k_i}$ as $y_i$ and $x_{k_j}$ as $x_j$ to ease the notational burden.

Each response is the result of a dyadic interaction between an event $(k_i, t_i)$ in the past and the potential occurrence of an event with type $k_j$ at the present $t_j$. The bank of $R$ basis functions combines Gaussian spatial proximities and exponential clocks with support forward in time: % (forwardly asymmetric in time)
\begin{align}
g_r(x_j,y_i)&\coloneqq(2\pi\beta_r^2)^{-\frac{1}{m}}\exp{\frac{\norm{x-y}^2}{2\beta_r^2}},\label{eq:g}\\
f_r(\tau)&\coloneqq\kappa_r\mathrm{e}^{-\kappa_r\tau}\label{eq:f}.
\end{align}

Eqs.~\ref{eq:g}~\&~\ref{eq:f} form the spatial and temporal basis of $r=1,2,\dots R$ kernels comprising the response function, tentatively produced below and expanded with full nuance in Eq.~\ref{eq:response_full}. Evidently, our construction enables multiscale responses in space (with spreads $\beta_r$) and time (with decays $\kappa_r$). Later we introduce the ingredients $\xi(k_i)$ and $\gamma_r$ to relieve a constraint on the magnitude of each response.
\begin{equation*}\label{eq:response}
h(k_j,k_i,\tau)=\vmathbb{1}[\tau>0]\sum_{r=1}^R g_r(x_j,y_i)f_r(\tau)
\end{equation*} % v complete
The proposed model family is framed as a strict superset of the parametric Hawkes process with exponential kernels: setting $R=1$, as we did in most of our experiments, collapses back to the exponential temporal response function.

A generalized Poisson process yields a clean log likelihood function, written as follows:
\begin{equation}\label{eq:likelihood}
\log L = \sum_{j=1}^N \log\lambda({k_j},t_j) - \sum_{k=1}^n\int_0^T \lambda(k,t)\mathrm{d}t.
\end{equation}
However, direct gradient-based optimization is prone to instability and a painstaking selection of hyperparameters. We turn to an \emph{E-M} procedure with an augmented objective.

%has proven unwieldy in anything other than deep general models. Even worse, its current form is not amenable to analysis.

Suppose we happened upon the expected \emph{branching structure}~\cite{ref:halpin,ref:reinhart} of the realized point process. In other words, we introduced latent variables $[p_{ijr}]\in\mathbb{R}^{N\times N\times R}$ holding expectation estimates of $P_{ijr}\in\{0,1\}$, indicating whether it was the event instance $i$
 that triggered instance $j$, and attributing responsibility to kernel basis $r$. Approximate knowledge of the untenable true line of causation endows us with the so-called complete-data log likelihood termed $\log L_c$~\cite{ref:yuan,ref:schoenberg,ref:reinhart}, an expectation of the joint log-probability density of the record and the latent variables $[P_{ijr}]$ in terms of their probabilities $[p_{ijr}]$.
\begin{multline}\label{eq:complete_likelihood} % A MIXTURE LANDSCAPE OF ATTENUATED GAUSSIANS OVER TIME, SAMPLED AT SPECIFIC LOCATIONS
\log L_c \coloneqq \sum_{j=1}^N\Bigg(\sum_{i=1}^N \sum_{r=1}^R p_{ijr}\log h_r(k_j,k_i,t_j-t_i) + p_{bj}\log\mu({k_j})\Bigg)- \sum_{k=1}^n\int_0^T \lambda(k,t)\mathrm{d}t
\end{multline}
By abuse of notation let $h_r(\cdots)$ denote the $r$th response kernel. Note that in the above form, what was previously a logarithm of summations (see Eq.~\ref{eq:likelihood} and~\ref{eq:intensity}) is replaced by a weighted sum of decoupled logarithms. The probability that the event instance was due to the background white Poisson process is $p_{bj}$; $\forall j,\ \sum_{i,r} p_{ijr} + p_{bj}=1$. Concretely, given a model $\lambda(k,t)$, our expectation for the particular branch of event cascades where event $i$ triggered event $j$ via kernel $r$ is the ratio of that particular contribution to the overall intensity:
\begin{equation}\label{eq:causalities}
p_{ijr}\coloneqq\frac{h_r(k_j,k_i,t_j-t_i)}{\lambda({k_j}, t_j)}.
\end{equation}

\begin{lemma}\label{prop:bound}
  The complete-data log likelihood in Eq.~\ref{eq:complete_likelihood} provides a lower bound for the point process log likelihood (Eq.~\ref{eq:likelihood}): $\log L_c \leq \log L.$ % minorizes the log likelihood?
\end{lemma} % maximize minorize algorithm?

\begin{proposition}\label{prop:general}
  %Maximizing $\log L_c$ along the model's parameters while holding $[p_{ijr}]$ fixed never decreases $\log L.$
  The lower bound of Lemma~\ref{prop:bound} is maintained when the latent variables' expectations $[p_{ijr}]$ take on any value as long as $\forall j,\ \sum_{i,r} p_{ijr} + p_{bj}=1.$
\end{proposition}
Our proof of Proposition~\ref{prop:general} relies on Lemma~\ref{prop:bound} in the supplementary material. The right-hand term in Eq.~\ref{eq:complete_likelihood} simplifies vastly if one assumes that $\forall y_i\in Y,\ \forall r,\ \sum_{x_k\in X} g_r(x_j, y_i)=1$, and that $\forall i\forall r,\ \int_{t_i}^T f_r(t-t_i)\mathrm{d}t\approx1$. The latter approximation is tenable for large enough $T$; the former is not. Only through certain concessions may we gain confidence that the sum is roughly unit (equal to one). First note that by the Gaussian integral, $\forall y\int_{\mathbb{R}^m}g_r(x,y)\mathrm{d}x=1$.
Veen \& Schoenberg~\citeyear{ref:schoenberg} approximated this sum as a Gaussian integral and demonstrated the viability of such alongside later studies. This approximation holds as long as the spatial occurrence of events is distributed uniformly in $\mathbb{R}^m$. In our embedding scheme, they are not: events are clumped at discrete locations of their types, and the objective function must also be constrained so that it does not drift into regimes that violate the approximation. Thusly, we are left with a coerced normalization of Eq.~\ref{eq:g}; % trivially satisfied
\begin{equation}\label{eq:new_g}
\hat g_r(x,y)=\frac{ g_r(x,y) }{g_r(y)},\quad g_r(y)\coloneqq \sum_{x'\in X} g_r(x',y).
\end{equation}
We rely on the unaltered form in Eq.~\ref{eq:g} for the derivation of analytical maximizers during optimization, as is well established in seismology and other spatiotemporal studies~\cite{ref:reinhart}; nevertheless, we found empirically that the intervention in Eq.~\ref{eq:new_g} stymies potential drift towards degeneracy. Granular control of the magnitudes is desirable, so the final touch is the introduction of one more kernel parameter, $\xi(k_i)$, to directly represent the total after summation. The spatial kernel thus handles only the distribution of influence across its receptors. The full response function is therefore Eq.~\ref{eq:response_full}. We eliminate redundancies by \textit{post hoc} constraining the \emph{exertion coefficient} $n^{-1}\sum_{l=1}^n \xi(l)=1$ and scaling the \emph{basis coefficients} $\gamma_r$ appropriately. % anecdotally rather than empirically
\begin{equation}\label{eq:response_full}
h(k_j,k_i,\tau)\coloneqq\vmathbb{1}[\tau>0]\sum_{r=1}^R \xi(k_i)\gamma_r \hat g_r(x_j,y_i)f_r(\tau)
\end{equation}

\subsection{Optimization}

Furnished with the expected branching structure in Eq.~\ref{eq:causalities} (the ``Expectation'' step), we perform projected gradient ascent by setting partial derivatives of the complete-data log-likelihood with respect to each kernel parameter to zero (the ``Maximization'' step). Eventually, the potential trigger routes $p_{ij}$ are aggregated in certain ways to form coefficient estimates. Omitting the domains of summation over $i$ and $j,i$ as implicitly $\{1,2,\dots N\}$ and $\{1,2,\dots N\}\times \{1,2,\dots N\}$ respectively, the solutions unfold as the following: \vspace{0.1em}\\
\begin{minipage}{.49\textwidth}
\begin{align}
\kappa_r(\alpha_\delta, \beta_\delta) &= \frac{\sum_{j,i} p_{ijr} + \alpha_\delta - 1}{\sum_{i,j}p_{ijr}(t_{k_j}-t_{k_i}) + \beta_\delta},\label{eq:delta_prior}\\
\beta_r^2 &= \frac{\sum_{i,j}p_{ijr}\norm{x_{k_j}-y_{k_i}}^2}{m\sum_{j,i} p_{ijr}},\label{eq:beta}\\
\gamma_r &= \frac{\sum_{j,i} p_{ijr}}{\sum_{i} \xi(k_i)},\label{eq:gamma}
\end{align}\end{minipage}
\begin{minipage}{.49\textwidth}
\begin{align}
\xi(l) &= \frac{\sum_{r=1}^R\sum_{j,i}\vmathbb{1}[k_i=l] p_{ijr}}{\sum_{r=1}^R\gamma_r\sum_i\vmathbb{1}[k_i=l]}\label{eq:xi},\\
\mu(k) &= T^{-1}\sum_j \vmathbb{1}[k_j=k]p_{bj}.
\end{align}\end{minipage}\vspace{1.5em}

At times, it is necessary to preserve focus on the acceptable time horizons for a particular domain. A %, in bias against ``degenerate'' ones. A
$\textrm{Gamma}(\alpha_\delta, \beta_\delta)$ prior on the decay rate $\kappa_r$ admits the maximization a posteriori in Eq.~\ref{eq:delta_prior}, which trivially becomes uninformative at the assignment $\kappa_r(1,0)$ that we chose in our upcoming experiments. We included those extra parameters simply in case it becomes desirable to bias the model's time scale in the future. Empirically, we found that one is typically interested in the half-life $(\log2/\kappa_r)$, the prior of which is the reciprocal of the aforementioned gamma distribution and characterized by the aptly named inverse-gamma distribution with expectation $\frac{\beta_\delta}{\alpha_\delta-1}$. Preserving the mean while increasing both parameters strengthens the prior.

The influences $\Phi=[\varphi(k,l)]$ consist of the kernels with time integrated out, i.e. %also known as adjacency, affinity, proximity, depending on the field's parlance
\begin{equation*}
\varphi(k,l)\coloneqq\int_0^\infty h(k,l,\tau)\mathrm{d}\tau=\sum_{r=1}^R \xi(l)\gamma_r \hat g_r(x_k,y_l).
\end{equation*}
There is evidence that this quantity encodes the causal network structure~\cite{ref:achab,ref:etesami}. Pursuant to the above maximization step, one may alternatively estimate all of these $n^2$ degrees of freedom: that is what we will later term our baseline.
%\begin{equation}\label{eq:phi}
%\varphi(k,l) = \frac{\sum_{j,i}\vmathbb{1}[k_j=k, k_i=l] \sum_{r=1}^R p_{ijr}}{\sum_i \vmathbb{1}[k_i=l]}.
%\end{equation} % NO \varphi_r here. That only serves to confuse.

\subsection{Embedding the Dyadic Relationships}\label{sec:embedding}
We present two candidate approaches for estimating optimal Euclidean embeddings of the Hidden Hawkes Geometry (hence \texttt{HHG}) in a multivariate point process. One is based on the gradients of $\log L_c$, and the other on a diffusion-maps heuristic. Finally, we present a full-rank baseline estimator derived from the same EM algorithm.

\subsubsection{Maximum Likelihood}\label{sec:fp}%(\texttt{HHG-A/B})
We learn embeddings directly via the EM objective function, $\log L_c$, rather than some heuristic.
Our approach updates both the reception and influence embedding, concurrently with the rest of the parameters, during the M-phase. The influence points maximize their EM objective at tractable solutions to a set of decoupled equations.

Observe the partial gradient with respect to an influence vector $y$, after expanding the response functions:
\begin{multline}\label{eq:partial_y}
\frac{\partial\!\log L_c}{\partial y_l}=\sum_{r=1}^R\Bigg[-\sum_{j,i}\vmathbb{1}[k_i=l]p_{ijr}\left(\frac{y_l-x_{k_j}}{\beta_r^2}\right)\\+\xi(l)\gamma_r\sum_i\sum_{k=1}^n\vmathbb{1}[k_i=l](2\pi\beta_r^2)^{-m/2}\left(\frac{y_l-x_k}{\beta_r^2}\right)\exp{-\frac{\norm{y_l-x_k}^2}{2\beta_r^2}}\Bigg]
\end{multline}
This expression is difficult to solve analytically. Recall, however, our prior simplifying assumption that $\forall y,r\ \sum_{x\in X} g_r(x, y)=1$, also enforced a posteriori by means of Eq.~\ref{eq:new_g}. See the discussion above that surrounds this equation. The latter portion of Eq.~\ref{eq:partial_y} contains the form $\sum_x (x-y) g_r(x, y)$, equivalent to taking a quantized ``expectation'' of a Gaussian variable subtracted by its own mean (see Eq.~\ref{eq:g}). %Hence a viable approximation to the solution stems from neglecting the contribution of
Hence, as long as the Gaussian sum is assumed to be approximately unit, then the entire second part of Eq.~\ref{eq:partial_y} vanishes due to the contribution of the pieces involving $x_k$. We garner the following---rather intuitive---formula for globally optimal influence points, within the microcosm of the current E-phase:
\begin{equation}\label{eq:y}
y_l=\frac{\sum_{r=1}^R\sum_{j,i}\vmathbb{1}[k_i=l]p_{ijr}x_{k_j}}{\sum_{r=1}^R\sum_{j,i}\vmathbb{1}[k_i=l]p_{ijr}}.
\end{equation}
Evidently each influence point $y_i\in Y$ is attracted to the reception points $\{x_j\in X\}$ that it appears to excite.

\paragraph{First-order optimizer (\texttt{HHG-A}).}

All parameters are updated simultaneously in each fixed-point iteration. Solving the optimality conditions leads to a decoupled system of equations comprising functionally distinct blocks of parameters, like reception points, influence points, and kernel decay rates. Unfortunately there is no analogous solution for the reception points that could manifest by a sensible approximation like that of the Gaussian integral, above. The simplest strategy is to submit to regular gradient ascent with learning rate $\varepsilon$: climbing the average log likelihood %for consistency across different record lengths:
\begin{equation}\label{eq:learning_rate}
x_k\leftarrow x_k+\varepsilon\cdot N^{-1}\frac{\partial\!\log L}{\partial x_k}.
\end{equation}
 %We also include the remark that classic $L_2$-regularization may suit applications in which the optimization landscape is not well behaved, thereby subjecting the latent space to a Gaussian prior.

We produce the gradient below, in implicit vector notation.
\begin{multline}\label{eq:partial_x}
a_k\coloneqq\frac{\partial\!\log L_c}{\partial x_k} = \sum_{r=1}^R\sum_i \left(\frac{x_k-y_{k_i}}{\beta_r^2}\right)\Bigg[-\sum_j\vmathbb{1}[k_j=k]p_{ijr}\\ + \xi(k_i)\gamma_r(2\pi\beta_r^2)^{-m/2}\exp{-\frac{\norm{x_k-y_{k_i}}^2}{2\beta_r^2}}\Bigg].
\end{multline}
To gain intuition on the selection of $\varepsilon$, we looked into \emph{entropic impact} as a heuristic. The beautiful findings of~\citeR{ref:mcfadden} allowed us to reason about the contribution of shifting an embedding point to the differential entropy of a doubly stochastic point process: % subsequent? forward in time?
\begin{equation*}
\frac{\partial^2 H(k,t)}{\partial t\partial x_k} = \left[\log\lambda(k,t) + 1\right]\sum_i \frac{\partial h(k,k_i,t-t_i)}{\partial x_k},
\end{equation*}
which admitted a simple rule of thumb for adjusting the learning rate of Eq.~\ref{eq:learning_rate} proportionally to $n/N$, with other factors pertaining to domain idiosyncrasies. Maintaining this rule ameliorated convergence in our synthetic experiments of \S\ref{sec:synthetic}.

\paragraph{Second-order optimizer (\texttt{HHG-B}).} % of second order?
Further differentiating Eq.~\ref{eq:partial_x} with respect to every reception point leads to a block-diagonal Hessian matrix, $B$, decoupled into $(m\times m)$ coordinate blocks indexed by event type, $B_k$. Assuming each of these is negative definite---more on that below---then it is numerically trivial to invert them.
A naïve Newton step could have trouble converging, so we employed Levenberg-Marquardt regularization to introduce a sort of learning rate, $\varepsilon_1$~\cite{ref:nesterov}.
Newton-Raphson optimization may be viewed as maximizing a downwards-facing (convex) local quadratic approximation of the objective function. From that perspective, $\varepsilon_1$ plays the role of inserting a parabola centered at the current $x_k$ and strengthening local convexity. We additionally introduce a regularization parameter $\varepsilon_2$ that constrains how far the embedding points may escape the origin. It may be useful in quelling the aimless drift of unconnected event types. Following the previous analogy, $\varepsilon_2$ superimposes a parabola at the embedding origin, a Gaussian well in the log-likelihood space. High $\varepsilon_1$ grows jump sizes to mimic a learning rate, whereas $\varepsilon_2$ shrinks the jumps as a regularizer.
\begin{align}\label{eq:newton_step}
  \tilde a_k \coloneqq a_k - 2(N\varepsilon_2)x_k,\quad
  \tilde B_k \coloneqq B_k - 2N(\varepsilon_1^{-1} + \varepsilon_2)I,\qquad
  x_k\leftarrow x_k+\tilde B_k^{-1}\tilde a_k.
\end{align}

In practice, we chose to execute four such optimization steps during each M-phase to approximately converge to the current global optimum. Recall that a sole global optimum exists for each block of parameters per M-phase; however, we are point-wise maximizing an upper bound to the true likelihood. Each EM iteration concludes at a solution that is not guaranteed to be optimal, much less global. Yet it does converge eventually~\cite{ref:lange}.

\newcommand{\diag}{\operatorname{diag}}

\paragraph{Convexity.} %Decompose Hessian, ensure convexity thereof, and relegate equations to the appendix? Or keep them all here? Together with regularization, we get convexity!
A multivariate function with negative-definite Hessian throughout the domain has one extremum, the global maximum, and is thereby convex. The reception-point Hessians $B_k$ are not negative definite per se. They may, however, be deconstructed into the difference of a diagonal component and an outer product of another matrix with itself: $B_k = \diag(c_k) - D_k^TD_k.$ Clearly, if the elements of $c_k$ were all negative, then the whole $B_k$ would be negative definite. The vector $c_k$ is %proportional to (up to a positive scalar)
\begin{equation*}
  k\mapsto\sum_{r=1}^R\beta_r^{-2}\sum_i\Bigg[\xi(k_i)\gamma_r(2\pi\beta_r^2)^{-m/2}\exp{-\frac{\norm{x_k-y_{k_i}}^2}{2\beta_r^2}} - \sum_j\vmathbb{1}[k_j=k]p_{ijr}\Bigg],
\end{equation*}
where the positive component inside the sum is bounded above by unit, and the negative component is not analogously bounded below. To avoid undesirable cases, we apply a zero ceiling to each element of $c_k$ before employing the additional regularizers via $(\varepsilon_1,\varepsilon_2)$. Convexity is therefore ensured. Oftentimes, it suffices to only set one of the two---particularly $\varepsilon_1$---to a nonzero value. Whenever only $\varepsilon_2$ is specified, it is implied that $\varepsilon_1\to\infty$.

\subsubsection{Diffusion-maps heuristic (\texttt{HHG-DM})}\label{sec:dm}
Could we posit a diffusion process across event types, and estimate coordinates that recreate diffusion distances on an approximate Riemannian manifold? Random-walk manifold embeddings are helpful in deep representations~\cite{ref:kalatzis,ref:li,ref:rey}. Construed as graph affinities, the influences $\Phi$ guide a Markovian random walk of which diffusion maps~\cite{ref:ghanem,ref:lian,ref:coifman} may be approximated via spectral decomposition. We found that asymmetrical diffusion-maps embeddings in the style of~\citeR{ref:pham} serve as an adequate initial condition for $(X, Y)$ but are not always conducive to stable learning in conjunction with the rest of our iterative procedure. We term the model learned entirely this way as \texttt{HHG-DM}.% and for brevity we relegate its review to Appendix B. %  in contrast to \texttt{FP} above and the upcoming baseline FRB.

We briefly review the technique's application here; the curious reader is encouraged to peruse the theory presented in Coifman and Lafon's seminal publication~\citeyear{ref:coifman}. Casting the influence matrix as edge weights in a bipartite graph flowing between $\textit{influence (col.)}\leftrightarrow\textit{reception (row)}$, we examine the diffusion process upon it~\cite{ref:pham}. We first normalize by density to our liking, per our selected value for the parameter $0\leq\alpha\leq 1$ according to
\begin{equation}\label{eq:dm_norm}
A = \diag\left(\big\langle\sum_{l=1}^n \varphi(l, k)\big\rangle_{k=1}^n\right)^{-\alpha} \cdot\  \scalebox{1.2}{$\Phi$}\ \cdot \diag\left(\big\langle\sum_{l=1}^n \varphi(k, l)\big\rangle_{k=1}^n\right)^{-\alpha}.
\end{equation}

Consider the row-stochastic version of $A$, named $B_R$. Its singular values multiplied by the left (orthonormal) eigenvectors thereof supply manifold embedding coordinates for the reception points, weighted by significance according to the singular values. Likewise, $B_I$ may be constructed as the row-stochastic transformation of $A^T$ from the same Eq.~\ref{eq:dm_norm}, of which the resultant coordinates grant us the influence points. In each set of coordinates, we preserve only those corresponding to the highest $m$ singular values---except for the largest, which is constant by definition.

\subsubsection{Full Rank Baseline (\texttt{FRB})}\label{sec:baseline}
Did the reduction in degrees of freedom lend its hand to a more generalizable model of the point process? In order to motivate the reason for having an embedding at all---besides the gains in interpretability---we pitted the techniques \texttt{HHG-A/B} and \texttt{HHG-DM} against the following:
\begin{equation}\label{eq:baseline}
h(k,l,\tau)=\varphi(k,l)\vmathbb{1}[\tau>0]\sum_{r=1}^R f_r(\tau),
\end{equation}
having estimated the full-rank matrix entries $\varphi(\cdot, \cdot)$ directly~\cite{ref:zhou2}.

\subsection{Initialization}
We surmised adequate initial conditions for the EM procedure with a fixed empirical protocol. The surmised influence matrix came by summing up correlations between event types. %according to an asymmetrical exponential kernel on their time separation.
\begin{equation}\label{eq:guess}
\hat \varphi(k,l) = \sum_{i,j} \vmathbb{1}[k_j=k, k_i=l]\cdot\hat\kappa\mathrm{e}^{-\hat\kappa\cdot(t_j-t_i)}
\end{equation}
Notice that it remains unscaled. Initial ``coefficient'' $\hat\kappa$ was computed as the naive reciprocal inter-arrival time between event types $\hat t = n(N-1)^{-1}\sum_{i=1}^{N-1} (t_{i+1}-t_i)$. We justify this construction on basis of Eq.~\ref{eq:delta_prior}, which forms a weighted average over said arrival times to garner an optimal estimate for $\kappa_r^{-1}$. We feed the result of Eq.~\ref{eq:guess} into the diffusion-maps algorithm in order to obtain our initial embeddings $(\hat x, \hat y)\in \hat X\times\hat Y$. $\forall r\hat\beta^2_r$ is initialized at the mean dyadic squared distance in the embedding; $\hat\kappa_r=(r\hat t)^{-1},$ in which variety is injected to nudge the kernels apart; $\hat\gamma_r=R^{-1}$ and finally $\forall x,\ \hat\mu(x)=T^{-1}n^{-1}N$. %We strive to minimize careless bias. % that is invariant to absolute scale by Eq.~\ref{eq:dm_norm}
%\subsection{EM-type Algorithm}

\section{Results}
First we explore simulated estimations to elucidate the comparative behavior of all four techniques: \texttt{HHG-A}, \texttt{HHG-B}, \texttt{HHG-DM}, and \texttt{FRB}. Then we present a number of findings and challenges regarding the characterization, namely, of a recent Ebola epidemic, the COVID-19 epidemic, and an illustrative portion of the options market. As the Euclidean norm, especially under a Gaussian kernel with exponential decay, suffers from the curse of dimensionality, all of the following experiments set $m\in\{2,3\}.$ It remains the topic of future study to incorporate parsimonious yet more robust metrics.

\subsection{Synthetic Experiments}\label{sec:synthetic}

We intended to stress-test the learning algorithm under small record sizes and numerous event types; we thereby contrived a few scenarios with known ground-truth parameters sampled randomly. The underlying models had a single kernel $R=1$ in the response function and conformed to the formulations from \S\ref{sec:embedding}. Each sampled model was simulated with the thinning algorithm (see e.g.~\citeR{ref:mei} and their supplementary material) in order to generate a time-series record of specified length $N$. Reception and influence points were realized uniformly from a unit square ($m=2$). Spatial bandwidths $\beta^2$ were granted a gamma distribution with shape $\alpha=\sfrac{1}{\sqrt n}$ and unit scale. Decay rates $\kappa$ were standard log-normal as were backgrounds $\mu(k)$, though scaled by $\sfrac{1}{n}$.

Stability~\cite{ref:bacry} was ensured by setting $\gamma=\sfrac{1}{\sqrt n}$, constraining the Frobenius norm $\norm{\Phi}_F= 1$ that upper-bounds the $L^2$-induced norm, which itself upper-bounds the spectral radius of the influences $\rho(\Phi)$, the real criterion.

\begin{table}[ht!]\centering
\begin{tabular}{l|ll|rrrr} % automatically generated from a pandas dataframe!
Settings & & & Outcomes & & \\
\toprule
   &    ($n$)   &   ($N$)   &  Test &  Train & Div. & Impr.\\
Model (see \S\ref{sec:embedding}) & {Types}  & {Events}  &                     &                &  & \\
\midrule
\texttt{HHG-B} ($\varepsilon_2=10^{-1}$,\hspace{-.5ex} {\small Eq.~\ref{eq:newton_step}})\hspace{-1.5ex} & 15 	& 300  &         $\boldsymbol{ -4.45 \pm 2.11}$ &
$-4.09 \pm 1.80$ & 0.15 & -0.023\\
\texttt{HHG-A} ($\varepsilon=0.75$,\hspace{-.5ex} {\small Eq.~\ref{eq:learning_rate}}) &  	&   &         $-4.82 \pm 2.80$ &
$-4.36 \pm 2.59$ & 0.14 & *0.028\\
\texttt{HHG-DM} 							 &  	&  &            $-5.21 \pm 4.33$ &
$-4.52 \pm 2.96$ & $\boldsymbol{0.12}$ & ***$\boldsymbol{0.040}$\\
\texttt{FRB}															 &  	&  &            $-5.34 \pm 3.88$ &
$-4.42 \pm 2.94$ & 0.20 & ---\\\specialrule{.03em}{.0em}{.1em}
\texttt{HHG-B} ($\varepsilon_2=10^0$)	 	 &     & 900 &             $-5.03 \pm 3.35$ &
$-4.79 \pm 3.88$ & 0.11 & 0.023\\
\texttt{HHG-A} 	($\varepsilon=0.25$)	 &     & &             $-5.79 \pm 3.98$ &
$-5.54 \pm 3.78$ & 0.12 & -0.012\\
\texttt{HHG-DM} 							 &   &  &            $\boldsymbol{-3.94 \pm 2.39}$ &
$-3.78 \pm 2.31$ & $\boldsymbol{0.08}$ & **$\boldsymbol{0.044}$\\
\texttt{FRB}							 &   &  &           $-6.13 \pm 3.61$ &
$-5.71 \pm 3.21$ & 0.14 & ---\\\specialrule{.03em}{.0em}{.1em}
\texttt{HHG-B} ($\varepsilon_2=10^0$)	 	 & 30  &    &             $-5.51 \pm 4.15$ &
$-5.47 \pm 4.16$ & 0.11 & 0.009\\
\texttt{HHG-A} 	($\varepsilon=0.5$)		 &     &    &           $\boldsymbol{-4.45 \pm 3.96}$ &
$-4.14 \pm 3.48$ & $\boldsymbol{0.09}$ & 0.011\\
\texttt{HHG-DM} 							 &   &  &           $-7.62 \pm 6.59$ &
$-7.04 \pm 5.92$ & 0.11 & *$\boldsymbol{0.020}$\\
\texttt{FRB}							 &   &  &           $-6.87 \pm 4.52$ &
$-6.46 \pm 4.17$ & 0.30 & ---\\
---& & & & & & \\
\texttt{HHG-B} ($m-1$)	 	 &     &    &             $-7.93 \pm 4.29$ & % ($m=1$)
$-8.21 \pm 4.78$ & 0.21 & -0.010\\
\texttt{HHG-B} ($m+1$)	 	 &     &    &             $-6.61 \pm 4.03$ &
$-7.04 \pm 4.29$ & 0.12 & -0.009\\
\texttt{HHG-B} ($m+2$)	 	 &     &    &             $-6.17 \pm 6.01$ &
$-6.60 \pm 6.28$ & 0.11 & $\boldsymbol{0.035}$\\
\texttt{HHG-B} ($m+3$)	 	 &     &    &             $-7.63 \pm 4.95$ &
$-7.92 \pm 5.47$ & 0.12 & 0.025\\\specialrule{.005em}{.0em}{.1em}
\bottomrule % quantiles for the highly skewed test-train gap?
\end{tabular}
\caption{\label{tab:synthetic} Agglomerate metrics collected in twenty simulated trials per synthetic experiment. %A line fit by total least squares revealed an approximate test/train ratio.
Outcome columns show, respectively, mean and standard deviation of test and train log likelihoods (cols. ``Test'' \& ``Train''), divergence from ground-truth branching structure (col. ``Div.''), and mean improvement in correlation of the model's embeddings from GloVe's, each with respect to the ground-truth space; t-test significance markers are provided as well (col. ``Impr.''). ***:~$P\leq0.01$, **:~$P\leq0.05$, *:~$P\leq 0.1$. The correlation improvements all passed an Anderson-Darling test~\cite{ref:razali} for normality with confidence $\geq 0.95$.}\vspace{0em} %Bold numerals indicate the sample rejected an Anderson-Darling test~\cite{ref:razali} for normality with significance $\leq0.05$.} % average is too loose of a term. ``on average'' is colloquial
\end{table} % estimated test/train ratio (col. ``ratio''), RMSE of the fit line (col. ``fit''),

\begin{figure}
\scalebox{0.9}{\input{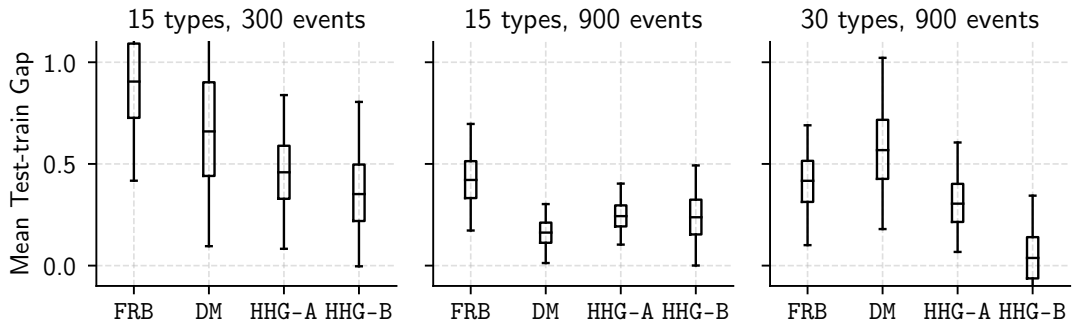}}\vspace{-1em}
\caption{\label{fig:synthetic-gaps} Decreases from in-sample (train) to out-of-sample (test) log likelihoods are indicative of overfitting. Bootstrapped means across the twenty trials reveal quantiles of 95\% confidence. Comparative behavior varies by $(n,N)$ configurations.}%\vspace{5em}
\end{figure}

In line with Goodhart's Law~\citeyear{ref:goodhart}, different facets of the model apparatus were scrutinized. First, we sought to ascertain whether the baseline tends to reach high in-sample likelihoods yet abysmal out-of-sample likelihoods, including extreme outliers. We bootstrapped the mean difference between the train and test log-likelihoods, arriving at confidence intervals for the test-train gap in Figure~\ref{fig:synthetic-gaps}. This metric is a common indicator of overfitting. Its applicability is more tenuous in the empirical records scrutinized in \S\ref{sec:real}, where the process is not guaranteed to be stationary.
%Specifically, not only are statistics on each likelihood important but also the relationship between the two. We thought to convey it through a total least squares~\cite{ref:groen} slope and centroid, followed by its root mean-square error (RMSE) in the first four outcome columns of Table~\ref{tab:synthetic}. and the slope how much the test varies with the train.

There are further questions one could ask than the descriptive means and standard deviations of train and test $\log L$ in Table~\ref{tab:synthetic}. Did our models recover the chain of causation between excited events? We opened up the empirical $[p_{ijr}]$ estimates and computed their Hellinger distance~\cite{ref:campbell} from those stipulated by the ground truth, as if they were categorical distributions. For each ``to be caused'' event $j$, the quantities $(i,r)\mapsto p_{ijr}$ form empirical histograms that are poorly suited numerically for the more conventional KL-divergence measure, unlike the Hellinger distance.

\begin{figure}
\scalebox{0.9}{\input{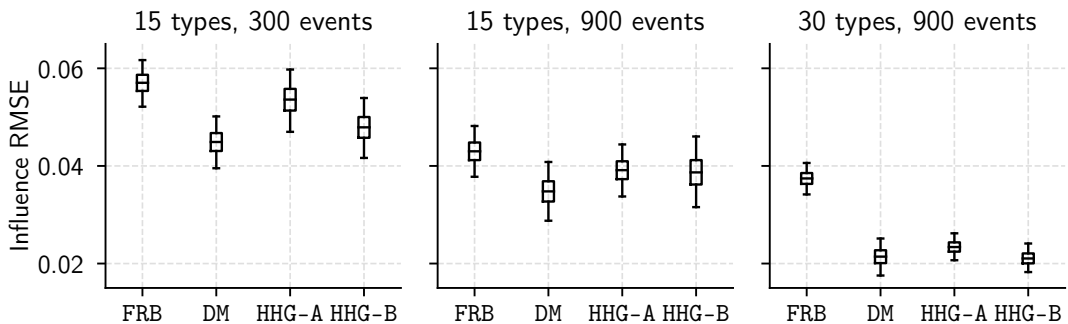}}\vspace{-1em}
\caption{\label{fig:synthetic-rmse} Root Mean Square Error (RMSE) of the recovered $(n\times n)$ influence (triggering) matrix, $[\varphi(k,l)]_{k,l}$. Bootstrapped means across the twenty trials reveal quantiles of 95\% confidence. Comparative behavior varies by $(n,N)$ configurations.}
\end{figure}

Was the matrix of asymmetric influences $\varphi(k,l)$ recovered correctly? We visualized the squared errors between the ground-truth influences and those of the final estimate in Figure~\ref{fig:synthetic-rmse}.

\begin{figure}
\scalebox{0.9}{\input{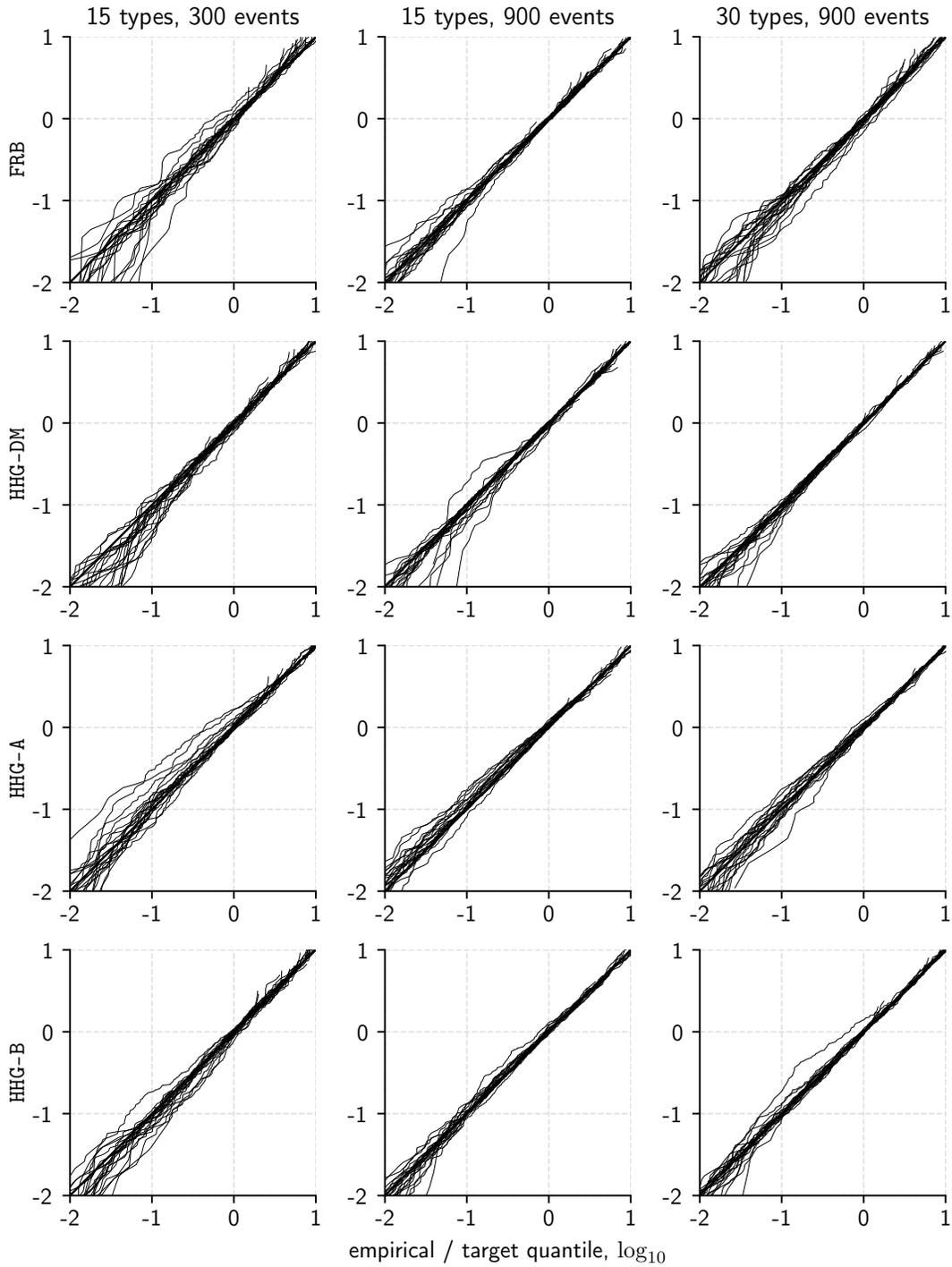}}
\caption{\label{fig:synthetic-quantiles} Synthetic results. Aligning quantiles of empirical and theoretical distributions of background-event interarrival times, deduced by the respective modalities. Deviations from the diagonal suggest a model mismatch, particularly in filtering out the background events. }
\end{figure}

Digging deeper, it is often pertinent to examine residual distributions. Our point-process model considers some events as (probably) excited by the past and others as purely white-noise background occurrences. These background events innovate the process by possibly triggering new dynamics. Should our model fit the record, then it may be called upon to sample the likely background events from the rest. The interarrival times of that subset ought to follow an exponential distribution with rate parameters $\mu(k)$. By the Poisson superposition principle~\cite{ref:ogata}, we may inspect the distribution of all background events against $\textrm{Exp}(\mu\coloneqq \sum_k \mu(k)).$ Those quantiles are illustrated in Figure~\ref{fig:synthetic-quantiles}.

Finally, as the intrinsic dimensionality $m$ is unknown in practice, we reevaluated \texttt{HHG-B} under mismatched dimensionalities against the static $(m=2)$ ground truth. These results were included for the $(30,900)-$case on the bottom of Table~\ref{tab:synthetic}. $(\varepsilon_1,\varepsilon_2)$ remained unchanged.

\paragraph{Comparison to GloVe.} In addition to excitatory point processes, our technique shares commonalities with a dual realm: that of vector embeddings for words and other sequential entities. We sought to draw comparisons with the latest treatments in that field. Our plan consisted of feeding the ordered sequence of event-type occurrences into a typical application of the GloVe scheme~\cite{ref:pennington}, in an attempt to recover a set of vectors that served as both influence and reception points. We designed a forward-looking window on the three next events to produce an asymmetric co-occurrence matrix, resulting in three embedding dimensions.

The final column of Table~\ref{tab:synthetic} displays a systematic evaluation against GloVe's embeddings with reference to the ground-truth geometry. Concretely, all pairwise distances in each setting (that of our learned model and the newfound GloVe embeddings) were correlated to those of the ground truth by Kendall's rank--based nonparametric statistic~\cite{ref:newson}. The gap between GloVe's estimated correlation and the model's under scrutiny, each in $[-1,1]$ and where a positive difference means the model correlated more with the ground truth, was collected in each trial, and means along with a t-test significance~\cite{ref:student} were reported in the last column of Table~\ref{tab:synthetic}. GloVe is nondeterministic, so we obtained the sample mean of ten embedding correlations per trial.%---a move that favors GloVe's results. %From the outcomes, it is evident that \texttt{HHG-DM} models pick up the spatial coordinates more consistently---probably due to the regularity imposed by their normalized spectral decomposition.

\subsection{Real-World Experiments}\label{sec:real}

\begin{table*}[ht!] % TODO model fits, even just of our own models
  \begin{tabular}{c c c c c | c c c}\toprule % test @ best train?
     \multicolumn{5}{c}{Model Variety (see \S\ref{sec:embedding})} & \multicolumn{3}{|c}{State of the Art, reported in~\citeR{ref:salehi}}\\
     \texttt{HHG-B} & \texttt{-A} & \texttt{-DM} & \texttt{FRB} & \texttt{GEO} & VI-SG~\citeyear{ref:salehi} & SGLP~\citeyear{ref:xu} & ADM4~\citeyear{ref:zhou1}  \\
    \midrule
    %Best Train $\log L$ & $0.50$ & $0.058$ & $0.87$ & $0.33$ & --- & --- & --- \\
     $-0.46$ & $-1.18$ & $-1.80$ & $-0.17$ & $-1.15$ & $-2.06$ & $-3.03$ & $-4.61$ \\
    %Half Life (days) & $8.02$ & $19.27$ & $7.51$ & $2.06$ & \multicolumn{2}{c}{\sl ---nonparametric---} &$6.93$ \\ \bottomrule\vspace{-.0em}\\
    %8k/17k & $-2.54$ & $-2.43$ & $-2.04$ & $-1.57$ \\
    %% JUST REMOVED THIS first 5k/17k & &$-3.05$ & $-2.44$ & $-3.37$ & $-1.42$ \\
    \bottomrule
  \end{tabular}
  \caption{\label{tab:ebola} Test $\log L$ at the best train $\log L$ under various conditions from the Ebola dataset. Each proposed model shows the best of 500 epochs; the others were trained until convergence. Log likelihoods are averaged over the record size. %Under $R=2$, half lives were computed numerically.
  \texttt{GEO} refers to interactions modeled over actual geographic coordinates. ($m=3,\ \varepsilon_1=10^{-2},\ \varepsilon_2=10^{-1},\ \varepsilon=2$)}
\end{table*}
\paragraph{Epidemics.}
Consider an infectious disease within a social apparatus, behaving like a diffusive point process. In 2019, a finely regularized variational approach to learning multivariate Hawkes processes from relatively short records~\cite{ref:salehi} was demonstrated on a dataset of symptom incidences during the $\sim$2014--2015 Ebola outbreak\footnote{\cite{ref:tini}}. We gave the record precisely the same treatment the authors did, and obtained significantly higher commensurate likelihoods than their best case. In turn, they outperformed the cutting-edge approaches MLE-SGLP~\cite{ref:xu} and AHHG-DM4~\cite{ref:zhou1} that regularize towards sparsity.

We also trained a model with an embedding fixed to the geographic coordinates of the 54 West African districts present in the dataset, requiring it only to learn the appropriate response function. Assessing the value of the real spatial component of this process, we found that only \texttt{HHG-B} and \texttt{FRB} successfully outperformed the spatiotemporal model. See Table~\ref{tab:ebola}. % Since our models \texttt{FP} \& \texttt{HHG-DM} are putatively data-efficient, we truncated the training set to its first 5,000 events and evaluated on the same test set as before, with a major temporal gap in between the two. See Table~\ref{tab:Ebola}; also, view the fruitful foray on two regional records of COVID-19 cases that follows. %foray into the COVID-19 pandemic and a fruitful result on South Korea. % analogous foray? purportedly
% Oddly, however, none of the scrutinized methods outperformed our baseline that directly learns the full-rank interaction matrix.

\paragraph{COVID-19.}

\begin{figure*}[ht!]\centering
\begin{minipage}{0.49\textwidth}
  \scalebox{0.78}{\input{korea-gpu-map.pgf}}
\end{minipage}\hfill
\begin{minipage}{0.49\textwidth}\hfill
\begin{tabular}{l l r}\toprule
  Model & Parameters & Test\\
  \midrule
  \texttt{HHG-B} & ($\varepsilon_1=10^2, m=2$) & $-3.05$\\
  \texttt{HHG-A} & ($\varepsilon=10, m=2$) & $\boldsymbol{-3.03}$\\
  \texttt{HHG-DM} & ($\alpha=1, m=2$) & $-3.28$\\
  \texttt{FRB} & & $-4.00$\\
  \texttt{GEO} & & $-3.11$\\
  \bottomrule
\end{tabular}
\end{minipage}
\caption{\label{fig:korea}\textbf{Left.} Geographic locations of South Korean regions (mostly cities). Colors were interpolated by hue on the basis of physical proximity to metropolitan Seoul (violet) versus Busan (red), the two major urban centers. \textbf{Right.} Out-of-sample average log-likelihoods attained by each model.}
\end{figure*}

\begin{figure*}[ht!]\centering
\scalebox{0.75}{\input{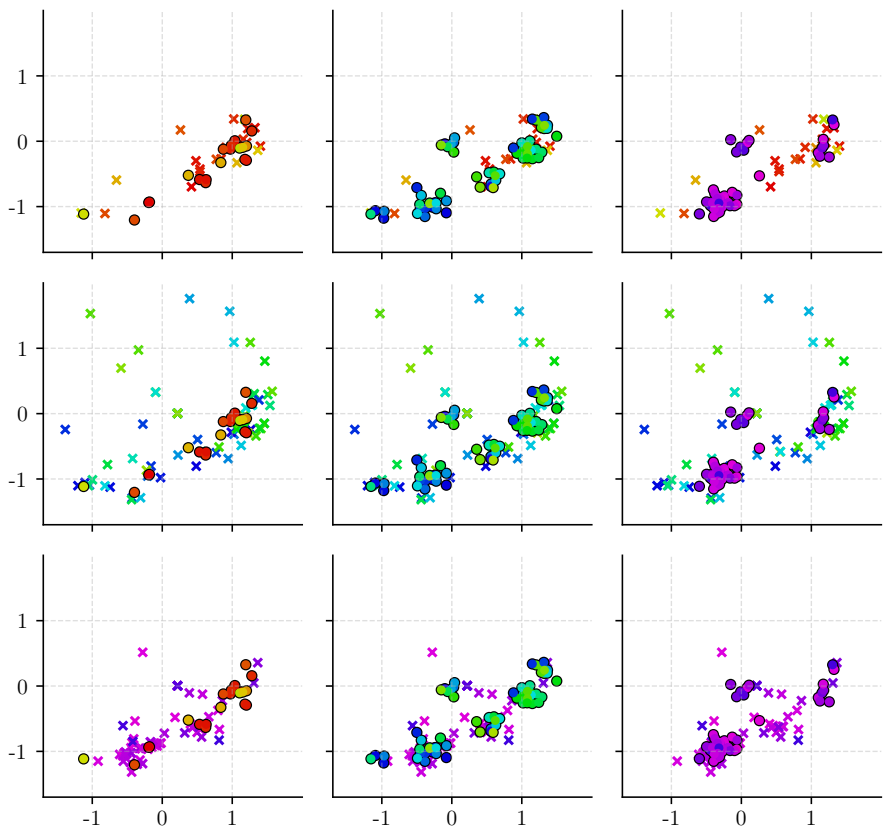}}
\caption{\label{fig:korea_embedding}Embeddings of the highest-scoring model for South Korea: our novel \texttt{HHG-A}. Each location is endowed with a pair of an ex and a dot, corresponding to receiving and influencing points respectively. Panels were faceted by qualitative color groups interpreted from Figure~\ref{fig:korea}. Under $m=2$, the raw hidden-space coordinates were plotted. We added normal noise with standard deviation of five units to each of the influence points (dots) to reveal the various colors stacked on top of each other.}\vspace{0em}
\end{figure*}

The novel coronavirus brought the world to its knees almost a year ago, as of this writing. The human suffering and socioeconomic disruptions have persisted since. Identifying harbingers for COVID-19's spatiotemporal transmission is of paramount importance for proactive policymaking. The most readily available data consist of confirmed cases tallied at the end of each day; timestamps of higher fidelity are both impractical and of questionable value, since a positive test result indicates that virus transmission occurred some days ago, the precise number of which is ambiguous. Further, the inhomogeneous testing rate per community and over time contributes to the nonstationarity of the process. Inherent nonstationarity stems from changing social practices and other exogenous conditions. Reports of deaths attributed to COVID-19 are potentially even less reliable as proxy indicators for transmission, due primarily to inconsistent protocols. As before, we only modeled point-process excitations. Incorporating a self-limiting aspect---as in the compartmental models~\cite<e.g.>{ref:bertozzi}---remains the topic of future study.

\paragraph{South Korea.} % KOREA IS MORE CONTROLLED
%Questioning whether it was the quality of the reported confirmed cases at fault,
Early on in the pandemic, South Korea experienced a transient surge in cases that it swiftly suppressed by strict social controls. We believed we were likely to succeed in interpreting that record as a relatively stable Hawkes process, as opposed to most other countries that did not enforce the same measures. We found the meticulous dataset released by the Korean Centers for Disease Control \& Prevention~\cite{ref:korea}. They detail 3,385 incidences from the early outbreaks across the 155 regions of the country, each with at least one infection occurrence. Diligent testing appears to contribute to the embedding's identifiability.

The test set consisted of the last 30 days in the record, containing 631 incidences. The table in Figure~\ref{fig:korea} displays the resultant score of each model. It is also worth noting that the full-rank baseline \texttt{FRB} registered a significantly higher in-sample likelihood than all the other models did, indicative of excess overfitting. See the full \texttt{HHG-A} embeddings in Figure~\ref{fig:korea_embedding}. The optimal response function (with one kernel, $R=1$) predicted that each contagious individual infected $\gamma=0.714$ others on average, with a half life of $3.34$ days.

% for posterity: LA was also done on the GPU.
\paragraph{Los Angeles} is a dense metropolitan area with diverse demographics. We are interested in recovering the landscape of transmission dynamics. It usually differs from a uniform spatiotemporal diffusion process because the physical separation between cities is not commensurate to how much their residents come into contact. Distant cities connected by common routes between home and work may be nearby in a putative ``interaction space.'' In the absence of reliable network information on contact rates---for mobility data introduce their own set of biases\footnote{\cite{ref:oliver}}---it would be useful to infer this from the infection rates themselves.

\begin{table*}[ht!]\centering % TODO \texttt{HHG-DM} is kind of like a random search...
%  \begin{minipage}{0.32\textwidth}\hfill
  \begin{tabular}{l c}\toprule
    Model & Test\\
    \midrule
    \texttt{HHG-B} ($\varepsilon_1=10^{-2},\,m=3$) & $0.694$\\
    \texttt{HHG-A} ($\varepsilon=20,\,m=2$) & $0.655$\\
    \texttt{HHG-DM}  ($\alpha=1,\,m=3$) & $\boldsymbol{0.701}$\\
    \texttt{FRB} & $0.659$\\
    \texttt{GEO} & $0.684$\\
    \bottomrule
  \end{tabular}
  \caption{\label{tab:la} Out-of-sample average log-likelihood for \textbf{Los Angeles} attained by each model.}
\end{table*}

\begin{figure*}[ht!]
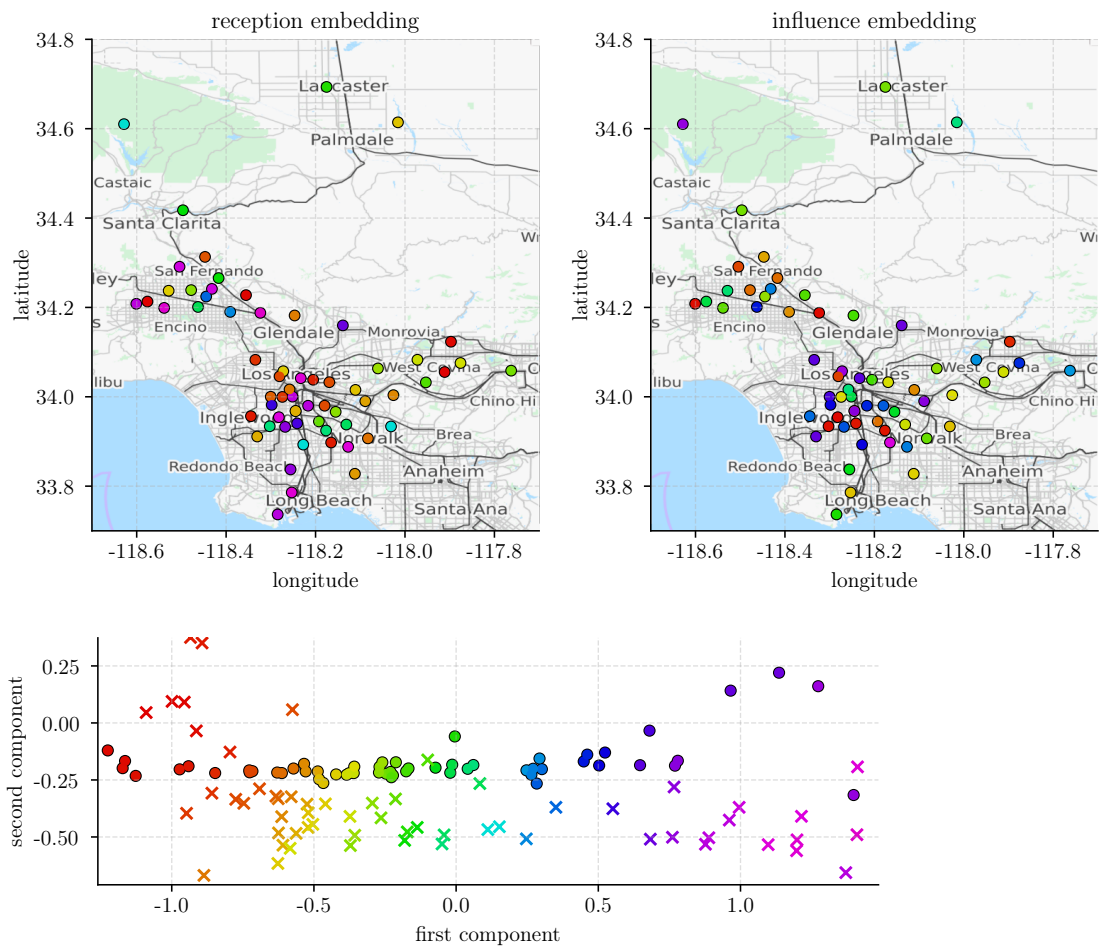

%  \end{minipage}
%  \begin{minipage}{0.65\textwidth}
    \scalebox{0.74}{\input{los-angeles-overlay.pgf}}\\
    \scalebox{0.74}{\input{los-angeles-embeddings.pgf}}
%  \end{minipage}
\caption{\label{fig:la}
\textbf{Above} are the geographic locations of the 64 cities in Los Angeles county that were impacted with the most confirmed cases of COVID-19. Communities are colored by locations in the reception (left) and influence (right) embedding in the highest-scoring \texttt{HHG-DM} model: from violet to red along the principal axis in the embedding space. Image source: OpenStreetMap.
\textbf{Below} are plotted the two principal components of the influence (dots) and reception (exes) embeddings accompanied by the same colorings as above.} %Source: ArcGIS accessed through GeoPy.
\end{figure*}

We found a dataset of confirmed cases in Los Angeles County curated by the Los Angeles Times~\cite{ref:latimes}. The sheer volume of infections per city is reflected in daily counts. To curb some of the nonstationary elements we truncated the record to the latest 50,000 cases as of December 12, 2020. To discretize the time series into events, we interpolated the cumulative daily counts of each city into a continuous signal. Specifically, we extended the curves linearly after a logarithmic (because of exponential growth) transformation. We then located the exact timestamp of consecutive increments of a specific threshold size; in this case study, that was ten infections. The resultant dataset consists of 5,000 events beginning roughly on November 9. Each of those events belongs to one of the 64 most impacted cities of Los Angeles County, depicted on the map in Figure~\ref{fig:la}.% and colored by a north-to-south gradient.

The test set consisted of the last five days in the record, comprising 1,605 events out of the 5,000. Trying $m=2,3$ and a handful of choices for the hyperparameter $\varepsilon,$ we arrived at the results in Table~\ref{tab:la}. Our novel parametrization \texttt{HHG-DM} scored the highest, and was visualized in Figure~\ref{fig:la} through colorings that reflect topography in the \texttt{HHG-DM} embedding. Its predicted average infection rate was $0.578$ and its half life was $0.0934$ days: a little troublesome in comparison to the South Korean results, and likely due to the instability of this process.

\begin{figure}[ht!]\centering
%\begin{minipage}[t]{0.45\textwidth}\centering
\scalebox{0.75}{\input{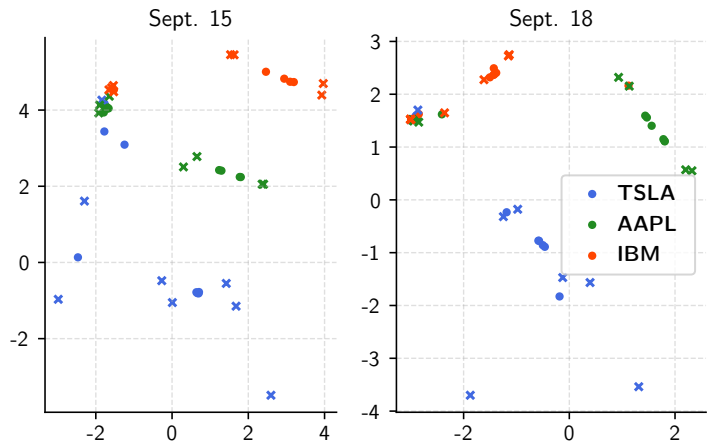}}\vspace{-.5em}
\caption{\label{fig:market-embeddings}\textbf{Embeddings} on market data. Eight trade partitions per stock; TSLA is blue, AAPL green, and IBM orange. Exes receive; dots influence. Embedding scales are normalized to reflect a unit kernel bandwidth.}
\end{figure}
\begin{figure}[ht!]\centering
\scalebox{0.75}{\input{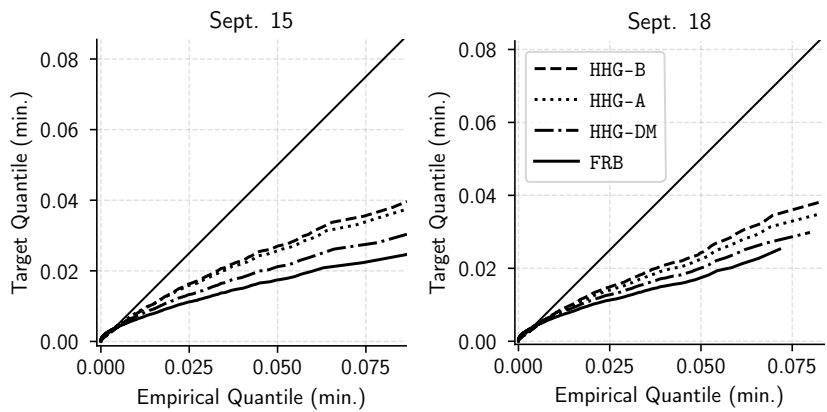}}\vspace{-.5em}
%\end{minipage}
\caption{\label{fig:market-quantiles}\textbf{Quantile-quantile plots} on inter-arrival times of all background market trades in the test partition, versus their supposed distributions from the sum of the model's background rates.}
\end{figure} % exogenous

\pagebreak

\begin{figure*}[ht!]\centering
  %\hfill\vspace{-0em}\\ what purpose did this line serve?
  \scalebox{0.82}{\input{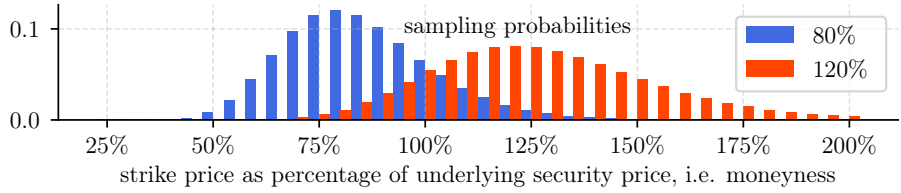}}\vspace{-.75em}
  \caption{\label{fig:options_sampling}Trades at discrete strike prices were resampled according to quantized log-Gaussian profiles with reference to moneyness at any given point in time. Standard deviations, in logarithmic space, were half the separation between the two densities' centers. They were kept ``loose'' for the sake of seamless translation even under abrupt fluctuations in the underlying stock price.} % They were carved into fuzzy partitions.
\end{figure*}

\paragraph{Options market.} The intertwined market activity of options with underlying stocks TSLA, AAPL, and IBM during the unremarkable consecutive trading days of Sept.\ 15 \& 18, 2017 led to 24 distinct event types. Roughly 100,000 total trades per day were sampled on 120\% and 80\% fuzzy moneyness levels---as portrayed in Figure~\ref{fig:options_sampling}---at the expiration dates 01/18/2019 \& 04/20/2018 for both puts and calls. The historical data was procured from AlgoSeek with the generosity of Prof.\ Roger G.\ Ghanem. The last 45 minutes of trading comprised each day's test set. See Table~\ref{tab:market}. % Roger G. Ghanem

An auxiliary accuracy metric in Table~\ref{tab:market} is derived from categorical cross-entropy of the predicted event type at the time of an actual occurrence, rendered by the expression $\exp{\mathbb{E}_i[\log\lambda(k_i,t_i) - \log\sum_l\lambda(l,t_i)]},$ in other words the geometric average of prediction accuracies. Between events, intensities decrease monotonically and with a homogeneous rate, so the ratio of intensities should stay roughly the same between occurrences---barring relatively outsized background rates. The ``naive'' score is computed from simply the mean empirical rate of each event type.
%%$\exp{\mathlarger{\mathbb{E}}_i\log\frac{\lambda(k_i,t_i)}{\sum_j\lambda(k_j,t_i)}}.$
%We visualized three-dimensional embeddings via their two principal components in Figure~\ref{fig:market}.%, and depict the cases with no time prior. %A Gamma$(32.1, 44)$ prior attempted to sway the estimators towards longer-term (on the order of seconds) behaviors, encouraging an expected half life of one minute.
We visualized the two-dimensional color-coded embeddings in Figure~\ref{fig:market-embeddings} and dsitributional fits for background events in Figure~\ref{fig:market-quantiles}.

\begin{table*}[ht!]\centering
\begin{tabular}{ll@{\hskip -.1ex}cccl}
\toprule
Dataset & Model & \multicolumn{1}{c}{Train} & \multicolumn{1}{c}{Test} & \multicolumn{1}{c}{Half Life (min.)} & Categorical Accuracy \\
%\multicolumn{1}{l}{Prior} &&& $(1,0)$ & $(32.1,44)$ & $(1,0)$ & $(32.1,44)$ & $(1,0)$ & $(32.1,44)$\\
\midrule
Sept.\ 15 & \texttt{HHG-B} ($\varepsilon_1=10^{1}$) & $\boldsymbol{3.13}$ & $\boldsymbol{2.60}$ & $6.50\times10^{-4}$ & $\boldsymbol{0.123}$ vs. naive\hfill$0.099$\\
          & \texttt{HHG-A} ($\varepsilon=20$) & $3.05$ & $2.57$ & $4.09\times10^{-4}$ & $0.119$ \hfill$0.099$\\
          & \texttt{HHG-DM} & $2.85$ & $2.34$ & $2.26\times10^{-4}$ & $0.103$ \hfill $0.099$\\
          & \texttt{FRB}  & $2.54$ & $2.11$ & $1.15\times10^{-3}$ & $0.103$ \hfill $0.099$\\
Sept.\ 18 & \texttt{HHG-B} ($\varepsilon_1=10^{-1}$) & $\boldsymbol{3.27}$ & $\boldsymbol{2.82}$ & $5.41\times10^{-4}$ & $\boldsymbol{0.135}$ vs. naive\hfill$0.108$\\
          & \texttt{HHG-A} ($\varepsilon=1$) & $3.13$ & 2.72 & $5.88\times10^{-4}$ & $0.123$ \hfill$0.108$\\
          & \texttt{HHG-DM} & $2.88$ & $2.48$ & $2.32\times10^{-4}$ & $0.103$ \hfill $0.108$\\
          & \texttt{FRB}  & $2.75$ & $2.43$ & $8.78\times10^{-4}$ & $0.117$ \hfill $0.108$\\
\bottomrule\vspace{-.75em}\\
\end{tabular}
\caption{\label{tab:market} Market fits with associated half lives. In \texttt{HHG-A}, the best in-sample $\varepsilon$ was picked out of a handful of candidates. A similar grid search was enacted on powers of $10$ for $\varepsilon_1$ in \texttt{HHG-B}. Outcome of 1,000 epochs depicted. Categorical accuracy is a prediction score $\in [0,1]$ for the next event type, at the point of occurrence of the next event.}
\end{table*}

\begin{figure}[!hb]
\scalebox{0.9}{\input{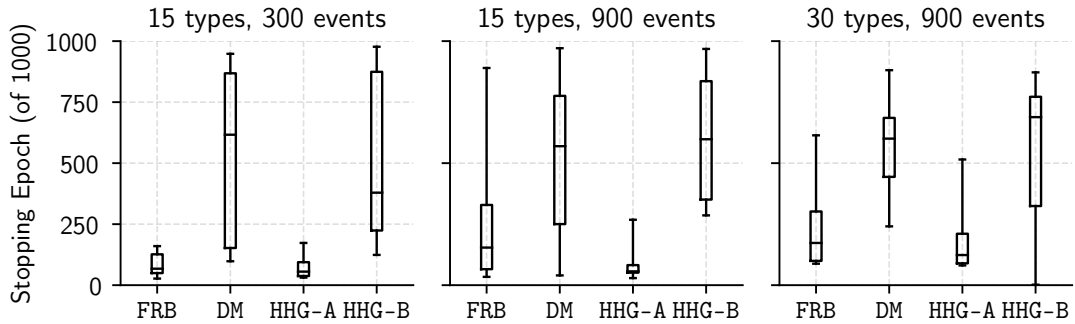}}\vspace{-1em}
\caption{\label{fig:synthetic-indices} Indices of the stopping epochs for the synthetic experiments. Closer to the end (1,000 epochs) suggests convergence, hence stability of the algorithm.}
\end{figure}

\section{Discussion}
We expose some empirical details for estimating models and then analyze significant results.

\subsection{Convergence}
Of our three proposals for hidden-embedding point process estimators \texttt{HHG-B}, \texttt{HHG-A}, and \texttt{HHG-DM}, we concede that the iterator's stability tends to decline in that order of enumeration.

Observe, in Figure~\ref{fig:synthetic-indices}, the epochs of the best models found. \texttt{HHG-DM} and \texttt{HHG-B}
converge best in the synthetic experiments. Interestingly, that tendency did not transfer onto real datasets. % bear out
That phenomenon demonstrates the value of theoretical guarantees for convergence.
\texttt{HHG-DM} appears to learn in the same way that a random search with some upward drift does, contrary to our initial expectations. % Monte Carlo sampler?
Every now and then the \texttt{HHG-DM} sampler would stumble upon a model with high in-sample likelihood that also garnered the best fit out of sample.

On the other hand, \texttt{HHG-A} and even more so \texttt{HHG-B} typically achieved their ideal fits at the end of their learning curves, even though \texttt{HHG-A} underwhelmed in the synthetic cases. In the Korean dataset, for instance, both \texttt{HHG-A} and \texttt{HHG-B} reached their maximum likelihoods at the last epoch. That was not the case for \texttt{HHG-DM}.

\subsection{Time-Scale Drift}
\begin{figure*}[ht!]\centering
  \scalebox{0.85}{\input{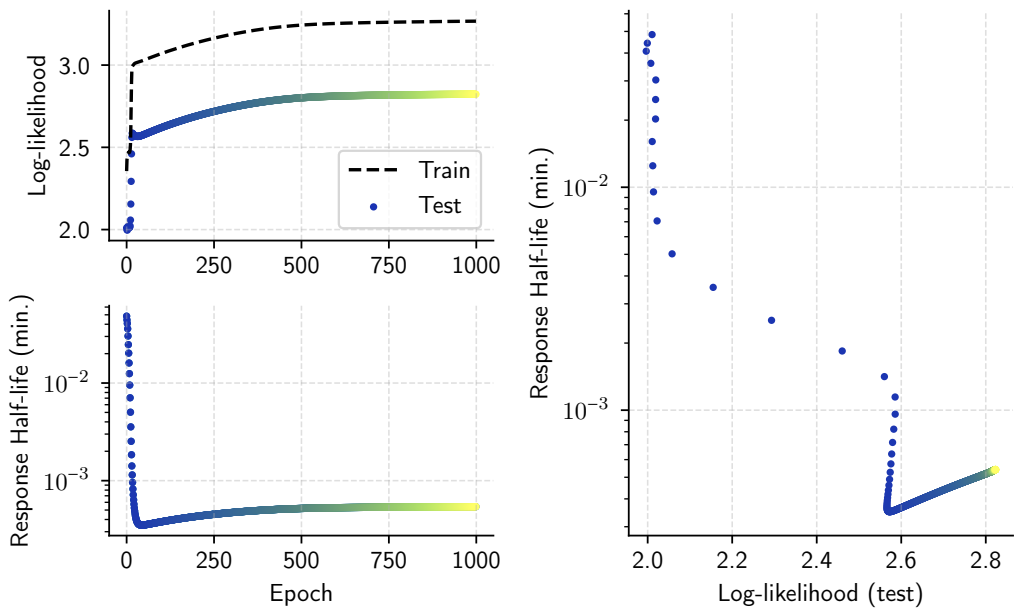}}\vspace{-.5em}
  \caption{\label{fig:options-drift} Illustrative scenario (Sept.\ 18 options) of time-scale drift in the model's response function alongside a growing test-set likelihood.}
\end{figure*}

As a model would converge to its optimal estimate, certain overarching parameters like the response time scale (homogeneous across event-type pairs) would drift.  See e.g.\ Figure~\ref{fig:options-drift}. It proved quite difficult to steer the optimizer away from its ``destined'' time scale.

\paragraph{Disentangling time scales.} Parsing out longer behaviors out of high-frequency trades is severely difficult, and the gamma prior failed to facilitate it. Whenever we implemented a prior strong enough to influence the learning algorithm, it would overpower the course of training and degenerate the model. Should one desire to investigate longer-term patterns through individual trades, it would be imperative to study higher-order interactions~\cite{ref:marmarelis} in the future.

\subsection{Predictive Ability}

%Our class of models (even ``FRB'') does not contend with overfitting as much as deep and expressive ones might. Summarily we do not have a validation set, and

We display the epoch with the best \emph{training} score in all of our experiments. %Most notable in Table~\ref{tab:synthetic} is how the \texttt{HHG-DM} formulation enjoys poorly determined systems, e.g.\ $(300, 90)$, but is typically outperformed on the basis of likelihoods by \texttt{FP} in better-posed situations like $(900,30).$ The baseline suffers in recovering actual causalities (measured by divergences from ground truth) in contrast to our novel models \texttt{FP} and HHG-DM.
Most notable in Table~\ref{tab:synthetic} of synthetic results are the following: %the test/train ratio, which measures how much the test $\log L$ falls for every unit of train $\log L$, is noticeably lowest in \texttt{FP} during the $(N=300, n=15)$ scenario and not very noteworthy anywhere else.
\texttt{HHG-A/B} yield a higher test $\log L$ than \texttt{FRB} in all three rows; \texttt{HHG-DM}'s results are a little more mixed. The ability to reveal chains of causality between excited events, as measured by the Hellinger divergence from the ground-truth branching structure, is better in \texttt{HHG} than in \texttt{FRB}. On average, the embeddings mostly mirror the ground-truth space more accurately than GloVe does, but \texttt{HHG-DM} here proved most consistent with significant improvement throughout.

We conjecture that the stiffness of a (normalized) spectral decomposition in \texttt{HHG-DM} provides some benefits and some drawbacks: it facilitates topological discovery---hence its correlation to the ground-truth layout---yet clashes with the real point-process objective. The direct optimizers of gradient-based \texttt{HHG-A} and curvature-informed \texttt{HHG-B} appear increasingly more resilient in general. % discernable embedding?

The excessive overfitting of the full-rank Hawkes process (\texttt{FRB}) is illuminated in Figure~\ref{fig:synthetic-gaps}, as well as the more granular Figure~\ref{fig:synthetic-quantiles}.

As is the case for convergence in the synthetic experiments, \texttt{HHG-B} and \texttt{HHG-DM} stand out as most accurately recovering the influence matrix in Figure~\ref{fig:synthetic-rmse}.

Focusing on the first two columns of the aforementioned figures, $(15,900)\ \&\ (30,900),$ we examine the relative performance of \texttt{HHG-B} and \texttt{FRB} as a function of increasing number of event types (the original process dimensionality). The full-rank baseline, \texttt{FRB}, exhibits inferior ability in recovering influences than \texttt{HHG-B} when the event types increase from $15$ to $30$, as witnessed in Figure~\ref{fig:synthetic-rmse}. A similar effect occurs to a lesser extent in Figure~\ref{fig:synthetic-quantiles}. The main driving force behind the failure of \texttt{FRB} appears to manifest in the over-determination of possible interactions between event types.

% MODEL MISSPECIFICATION COMMENT
Included additionally are the results of \textbf{model misspecification} for \texttt{HHG-B} in the $(30,900)$ case within the final rows of Table~\ref{tab:synthetic}. Standing against the two-dimensional ground truth, models with $m=1,3,4,5$ showcase the general sensitivity to over-compression ($m=1$) yet \texttt{FRB}-like performance following under-compression ($m=3,4,5$). A surprising outcome is the divergence and GloVe-correlation improvement of the under-compressed models. All three of those models exhibit consistent lowered divergence metrics. The experiments with $m=4,5$ yielded astounding correlation improvements, albeit of high variance. Sometimes---although not always, as is evident in the Ebola investigation below---a few extra degrees of freedom garner the helpful kind of flexibility.%, of the helpful variety. % sort/kind

\paragraph{Benchmark on the Ebola dataset.} Table~\ref{tab:ebola}'s results depict superiority in our models to the SotA. It is puzzling that our simple full-rank baseline EM estimator \texttt{FRB} outperforms the more sophisticated formulations, when MLE-SGLP resembles a regularized version thereof. Perhaps its shrinkage priors do not suit this particular dataset, or all $\sim$17,000 training events sufficiently fleshed out the possible interactions without need for regularization. Sparsity per se did not bear out as an appropriate constraint. Our hidden-embedding model \texttt{HHG-B} was the only such estimator that outperformed the spatiotemporal model built on geographic ``ground truths.'' This finding suggests that geographic information is recovered in that embedding, at least.

\begin{figure*}[ht!]\centering
  \scalebox{0.87}{\input{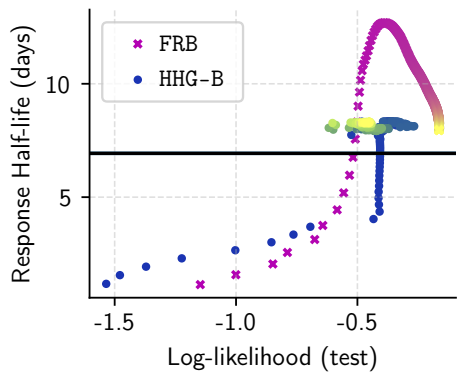}}\vspace{-.5em}
  \caption{\label{fig:ebola-paths} Per-epoch evolution of model estimates for the Ebola dataset. Black line indicates the decay-rate parameter set a priori for the other SotA models in Table~\ref{tab:ebola}.}
\end{figure*}
% Bring in mention of (fixed) decay rates. completely different problem framing. flexibility is good for our models??

It was difficult to transparently compare our methodologies against the SotA from \citeR{ref:salehi}. Part of the separation of outcomes could have been due to our framing that compels the estimator to learn a response decay rate on its own. The other cited models had those parameters selected via grid search, and remain static during the course of training. As witnessed in Figure~\ref{fig:ebola-paths}, our proposed models roughly converged to similar decay rates as those chosen in \citeR{ref:salehi}, but the paths followed were rather variable. % unfair comparison?

% SADLY, REMOVED THIS RESULT
%Rather than interaction sparsity, our models demonstrate that an overlaid Euclidean structure is especially more appropriate when the record is truncated and a full-rank estimator falters, as is evident in the second row of the table. There, \texttt{FP} \& \texttt{HHG-DM} suffer less than FRB and remain competitive to the SotA applied on the full training record, while using 29\% of the data. %It also seems that our models (typically with higher half lives) match the temporal scales of Ebola more closely, since the particular strain under consideration had an incubation period of about 8--12 days~\cite{ref:kerkhove}. % reflect reality

The half lives of all models are largely similar, aligning with the time scales of the particular Ebola strain under consideration that had an incubation period of about 8--12 days~\cite{ref:kerkhove}. % ballpark

% TODO we don't get as stable learning as FRB when m->n
\paragraph{Intrinsic dimensionality of the Ebola process.} One would expect that a model like \texttt{HHG-B} would replicate \texttt{FRB} as the degrees of freedom equilibrate, i.e.\ $m\to n$. However, that was not found to be the case. It appears that our fixed-point EM algorithms vastly complicate learning, for little payoff, in comparison to \texttt{FRB} when $m$ increases. We deemed it sufficient to demonstrate that the likelihood is not monotonic in embedding dimensionality $m$. To make our case, listed are likelihoods of the most stable \texttt{HHG-B} as a function of $m$: $\texttt{HHG-B}(2)=-1.7,\ \texttt{HHG-B}(3)=-0.46,\ \texttt{HHG-B}(4)=-0.80.$ It therefore appears that the optimal dimensionality lies at $m=3$, hinting at an intrinsic Euclidean geometry of the process.

\paragraph{Transmission network implicit in COVID-19 infection rates.} The manifold structure of contagion networks has long been studied~\cite{ref:taylor}, although estimation thereof remains elusive. Our point-process models identified embedding candidates based on the multivariate statistics of the infections. For one, we observe that the two urban hubs Seoul and Busan are not as far apart in the hidden influence space (Figure~\ref{fig:korea_embedding}) as they are geographically. They still cluster in their respective neighborhoods---Busan to the right, Seoul to the left. The generally rural regions in between seem to spread across this topology. Regarding Los Angeles (Figure~\ref{fig:la},) we first note that there is visible geographic locality preserved in the colorings. The reception embedding exhibits contiguous streaks of red/orange, of yellow/green, and of blue/violet. The influence embedding has communities of green/yellow scattered about. % likewise for influence embedding

There are clear and acute differences between the embeddings and geography, albeit some geographic structure is recovered in the embedding. We expected divergences, as previously stated, since physical communities and even mobility networks do not always correspond to the underlying transmission process. All these factors may serve to inform an epidemiological model as proxies, in agglomeration, to the true contact rates between individuals. We provide an additional factor, derived from the measured infection rates themselves. It is distinct from the other datasets that estimate interaction between communities. We also demonstrated its predictive ability.
%we note that the Lancaster and San Fernando valley embeddings are in close proximity. Their influence points appear almost mutually exclusive to the blue locations, corresponding roughly to central Los Angeles. Perhaps they follow distinct underlying processes? Southern Los Angeles, including Long Beach, overlaps with those northern cities more in the latent space.

On our results pertaining to the pandemic, we decided against including the auxiliary goodness-of-fit tests on inter-arrival times and classification accuracies. Our reasoning is that the COVID-19 infection process is clearly not a Hawkes process; it is unstable, multiphasic, and multifaceted. Our models provide a coarse approximation of its dynamics. Other quantities more specific to a Hawkes process may be misleading. On the other hand, the likelihood is salient on any point process and embeddings should yield qualitative meaning. Our other point-process applications demonstrate the validity of the presented models.

\paragraph{Analysis of market embeddings.} Events belonging to each stock ticker tend to attract influencing points of the same color in Figure~\ref{fig:market-embeddings}, with deviations probably due to relative perceptions by the market. Efficient estimators are necessary in order to discern a lack of stationarity; in this case, transferring our Sept.\ 15 \texttt{HHG-A} model onto the Sept.\ 18 test set yielded an average $\log L$ improvement from $2.57$ to $2.82$. Vice versa, transferring from Sept.\ 18 to Sept.\ 15 led to a decrease from $2.72$ to $2.45$. Thus point-process behavior largely persists, but not entirely. % neither significant, sizable, noteworthy, nor marginal
Further, aggregate $\xi(l)$'s confirm our broad intuition that out-of-money trades move markets the most~\cite{ref:yang}. % prevailing view

Quantile plots in Figure~\ref{fig:market-quantiles} indicate that the embedding models---all with the \texttt{HHG-} suffix---outperformed the baseline in statistically filtering out white background events according to their estimated probabilities $p_{bj}$. A constant-intensity Poisson process would witness arrival times distributed exponentially~\cite{ref:bacry}; therefore, any deviation from that in a model's sampled background events suggests the presence of uncaptured interactions.

We observed that the categorical accuracy measured in Table~\ref{tab:market}, assessing classification ability towards the next event to come by ignoring temporal granularities, is highest for \texttt{HHG-B} in the two trading days studied. It beats the naive score by at least $2.4\%$ each time.

\subsection{Modality Prescription}
Given the most consistently smooth learning curves under \texttt{HHG-B} during both synthetic experimentation and empirical investigations, we recommend the implementation of \texttt{HHG-B} in future endeavors. The final test-set likelihoods of \texttt{HHG-B} were always remarkably close to the top, if not the actually the highest. There was no empirical dataset scrutinized for which \texttt{HHG-B} failed significantly in comparison to \texttt{HHG-A}, \texttt{HHG-B}, or even \texttt{FRB}.

\section{Related Work}
The fundamental notion driving the doubly stochastic process, first attributed to Cox~\citeyear{ref:cox}, manifests in a variety of ways including the Latent Point Process Allocation~\cite{ref:lloyd} model. In fact, note the meteoric diversity of Bayesian methods permeating serially dependent point processes~\cite<e.g.>{ref:apostolopoulou,ref:detommaso}. Zhang et al.~\citeyear{ref:zhang} sample the branching structure in order to infer the Gaussian process (GP) that constitutes the influence function. GPs, usually accompanied by inducing points, sometimes directly modulate the intensity~\cite{ref:liu,ref:lloyd,ref:aglietti,ref:ding,ref:flaxman}. Linderman and Adams~\cite{ref:linderman} took an approach that estimated a matrix very similar to our $[\varphi_{ijr}]$, relying on discretized binning and variational approximation. Salehi et al.~\citeyear{ref:salehi} exploited a reparametrization trick akin to those in variational autoencoders in order to efficiently estimate the tensor of basis-function coefficients. %Apostolopoulou et al.~\citeyear{ref:apostolopoulou} conceived a nonlinear term that enables the full inference of a model with inhibitions. %A refreshing perspective on each pairwise interaction as its own process can be found in~\cite{ref:cho}. % unorthodox

Recent progress has been made in factorizing interactions with a direct focus on scalability~\cite{ref:nickel}, improving on prior work in low-rank processes~\cite{ref:lemonnier}. Block models on observed interaction pairs also exist~\cite{ref:junuthula}. While these all achieve compact Hawkes processes, our methodology distinguishes itself by learning a Euclidean embedding with the semantics of a metric space, not a low-rank projection.

Deep nonparametric (Poisson or even intensity-free) point process models\footnote{\cite<notably>[]{ref:mei,ref:lu,ref:ganem,ref:okawa,ref:omi,ref:sharma,ref:shang,ref:he,ref:du,ref:jia,ref:mehrasa}} made massive strides in the past few years. However, not all domains permit such expressive characterization. Our baseline estimator resembles most closely the one described in the work of Zhou et al.~\citeyear{ref:zhou2}, whereas the spatiotemporal aspect is inspired from the likes of Schoenberg et al.~\cite<see>[]{ref:schoenberg,ref:mohler,ref:yuan}. Variational substitutes in the EM algorithm have also been explored~\cite{ref:cho}. A concurrent study to ours by Zhu et al.~\citeyear{ref:zhu} parametrizes a heterogeneous kernel in real Euclidean space by deep neural networks.

\section{Conclusions}
We demonstrated the viability of estimating embeddings for events in an interpretable metric space by tying it to a self-exciting point process. The three proposed expectation-maximization algorithms extract parsimonious serial dependencies to different degrees of reliability, partly depending on record length and dimensionality. The optimizer of second order \texttt{HHG-B} converges most consistently to an accurate model. In comparison to learning the full triggering matrix (with \texttt{FRB},) our embedding models achieve a higher likelihood out of sample. The estimated time scale for Ebola contagion settles close to its natural rate. Our framework paves the way for generalization, extension to more elaborate models, and consequent potential for actionable insights.

\onecolumn
\section*{Appendix A}

\subsection*{Proof of Lemma 1}
We seek to demonstrate that $\log L_c \leq \log L$ always. Recall the likelihood function and its complete-data relative:
\begin{align*}
\log L &= \sum_{j=1}^N \log\lambda(k_j,t_j) - \Lambda, \\
\log L_c &= \sum_{j=1}^N\left(\sum_{i=1}^N \sum_{r=1}^R p_{ijr}\log h_r(k_j,k_i,t_j-t_i) + p_{bj}\log\mu({k_j})\right)- \Lambda,\qquad
\textrm{where }p_{ijr}=\frac{h_r(k_j,k_i,t_j-t_i)}{\lambda({k_j}, t_j)}.
\end{align*}

The intensity may be written as $\lambda(k,t)=\sum_{i=1}^N\sum_{r=1}^R h_r(k, {k_i}, t - t_i) + \mu(k)$. Compare each pairing of a receiving event's (indexed by $j$) additive contribution to the two objectives; on one hand, where $\log L$ contains $\log\lambda(k_j, t_j)$, $\log L_c$ reckons with
\begin{multline*}
\sum_{i=1}^N \sum_{r=1}^R p_{ijr}\log h_r(k_j,k_i,t_j-t_i) + p_{bj}\log\mu({k_j})\\
= \frac{\sum_{i=1}^N \sum_{r=1}^R h_r(k_j,k_i,t_j-t_i)\log h_r(k_j,k_i,t_j-t_i) + \mu(k_j)\log\mu({k_j})}{\lambda (k_j, t_j)}\\
= \frac{\sum_{i=1}^N \sum_{r=1}^R h_r(k_j,k_i,t_j-t_i)\log h_r(k_j,k_i,t_j-t_i) + \mu(k_j)\log\mu({k_j})}{\sum_{i=1}^N \sum_{r=1}^R h_r(k_j,k_i,t_j-t_i) + \mu(k_j)}.
\end{multline*}

To minimize distractions, we focus on the structure of what is found inside the logarithms and what is found on the outside. Accordingly, the above expression takes the abstract form $\frac{\sum_q w_q \log x_q}{\sum_q w_q}$ in terms of positive (for they serve as arguments to a logarithm) weights $w_q$ and contents $x_q$, over arbitrary indices $q$. The concave logarithm admits the following relation, by Jensen's inequality:
\begin{equation*}
\frac{\sum_q w_q \log x_q}{\sum_q w_q} \leq \log \frac{\sum_q w_q x_q}{\sum_q w_q}.
\end{equation*}
Since $w_q=x_q$ corresponding to expansion terms in the intensity function,
\begin{equation*}
\log \frac{\sum_q w_q x_q}{\sum_q w_q} = \log \left[\sum_{i=1}^N \sum_{r=1}^R h_r^2(k_j,k_i,t_j-t_i) + \mu^2(k_i)\right] - \log \left[\sum_{i=1}^N \sum_{r=1}^R h_r(k_j,k_i,t_j-t_i) + \mu(k_j)\right].
\end{equation*}
We can likewise compare $\log \frac{\sum_q w_q x_q}{\sum_q w_q}$ to $\log \sum_q x_q;$
\begin{align*}
\log \frac{\sum_q w_q x_q}{\sum_q w_q} - \log \sum_q x_q &= \log \left[\sum_{i=1}^N \sum_{r=1}^R h_r^2(k_j,k_i,t_j-t_i) + \mu^2(k_i)\right] - 2\log \left[\sum_{i=1}^N \sum_{r=1}^R h_r(k_j,k_i,t_j-t_i) + \mu(k_j)\right]\\
&= \log \sum_q x_q^2 - \log \Big(\sum_q x_q \Big)^2 \quad \leq 0 \quad\textrm{ by the multinomial theorem.}
\end{align*}

Hence
\begin{align*}
\frac{\sum_q w_q \log x_q}{\sum_q w_q} \leq \log &\frac{\sum_q w_q x_q}{\sum_q w_q} \leq \log \sum_q x_q.\\
\therefore \frac{\sum_{i=1}^N \sum_{r=1}^R h_r(k_j,k_i,t_j-t_i)\log h_r(k_j,k_i,t_j-t_i) + \mu(k_j)\log\mu({k_j})}{\sum_{i=1}^N \sum_{r=1}^R h_r(k_j,k_i,t_j-t_i) + \mu(k_j)} &\leq \log \Big[\sum_{i=1}^N \sum_{r=1}^R h_r(k_j,k_i,t_j-t_i) + \mu({k_j})\Big]\\
\sum_{i=1}^N \sum_{r=1}^R p_{ijr}\log h_r(k_j,k_i,t_j-t_i) + p_{bj}\log\mu({k_j}) &\leq \log \Big[\sum_{i=1}^N \sum_{r=1}^R h_r(k_j,k_i,t_j-t_i) + \mu({k_j})\Big].
\end{align*}

Therefore we hold that $\log L_c \leq \log L.$\qed

\subsection*{Proof of Proposition 2}
We are once again concerned with the behavior of
\begin{equation*}
j \mapsto \sum_{i=1}^N \sum_{r=1}^R p_{ijr}\log h_r(k_j,k_i,t_j-t_i) + p_{bj}\log\mu({k_j}),
\end{equation*}
now in the case that $p'_{ijr}=\frac{h'_r(k_j,k_i,t_j-t_i)}{\lambda'({k_j}, t_j)}$ and the contents of the aforementioned logarithms do not match---hence the apostrophes. Note that the above is equivalent to
\begin{equation*}
\sum_{i=1}^N \sum_{r=1}^R p_{ijr}\log \frac{h_r(k_j,k_i,t_j-t_i)}{\lambda({k_j}, t_j)} + p_{bj}\log\frac{\mu({k_j})}{\lambda(k_j, t_j)} + \log\lambda(k_j, t_j),
\end{equation*}
since $\sum_{i=1}^N \sum_{r=1}^R p_{ijr} + p_{bj}=1\ \forall j.$ Now let us examine the effects of a mismatch, $h_r' \neq h_r \iff p'_{ijr} \neq p_{ijr}.$ We have, essentially,
\begin{equation*}
\sum_{i=1}^N \sum_{r=1}^R p'_{ijr}\log p_{ijr} + p'_{bj}\log p'_{bj} + \log\lambda(k_j, t_j).
\end{equation*}
When $[p]$ and $[p']$ are put under the lens of discrete/categorical distributions, we may parse the expression in terms of a cross entropy
\begin{equation*}
-H([p'], [p]) + \log\lambda(k_j, t_j).
\end{equation*}
Since $[p']$ is independent of $\lambda(k_j, t_j),$ our quantity reaches its maximum when the distribution $[p']$ is set to equal $[p]$ (where $H([p'], [p])=H([p']) + D_{KL}([p']||[p])$ is minimized). The same cannot be said about maximization with respect to $[p],$ which depends on $\log\lambda(k_j, t_j)$ as well. The latter task proves useful in learning.

Thus when we perform a single learning iteration and vary the model parameters to evaluate $\log L_c$ on some $[p]$ fixed a priori, it will still offer a lower bound to $\log L.$\qed

\bibliographystyle{theapa} % theapa as a bibliography style doesn't work at all
\bibliography{refs}

\begin{thebibliography}{}

\bibitem[\protect\BCAY{Achab, Bacry, Gaiffas, Mastromatteo,\ \BBA\ Muzy}{Achab
  et~al.}{2017}]{ref:achab}
Achab, M., Bacry, E., Gaiffas, S., Mastromatteo, I., \BBA\ Muzy, J.-F.
  \BBOP2017\BBCP.
\newblock \BBOQ Uncovering causality from multivariate hawkes integrated
  cumulants\BBCQ\
\newblock {\Bem International Conference on Machine Learning}, {\Bem 34}.

\bibitem[\protect\BCAY{Aglietti, Bonilla, Damoulas,\ \BBA\ Cripps}{Aglietti
  et~al.}{2019}]{ref:aglietti}
Aglietti, V., Bonilla, E.~V., Damoulas, T., \BBA\ Cripps, S. \BBOP2019\BBCP.
\newblock \BBOQ Structured variational inference in continuous cox process
  models\BBCQ\
\newblock {\Bem 33rd Conference on Neural Information Processing Systems}.

\bibitem[\protect\BCAY{Apostolopoulou, Linderman, Miller,\ \BBA\
  Dubrawski}{Apostolopoulou et~al.}{2019}]{ref:apostolopoulou}
Apostolopoulou, I., Linderman, S., Miller, K., \BBA\ Dubrawski, A.
  \BBOP2019\BBCP.
\newblock \BBOQ Mutually regressive point processes\BBCQ\
\newblock {\Bem 33rd Conference on Neural Information Processing Systems}.

\bibitem[\protect\BCAY{Bacry\ \BBA\ Muzy}{Bacry\ \BBA\ Muzy}{2016}]{ref:bacry}
Bacry, E.\BBACOMMA\  \BBA\ Muzy, J.~F. \BBOP2016\BBCP.
\newblock \BBOQ First- and second-order statistics characterization of hawkes
  processes and non-parametric estimation\BBCQ\
\newblock {\Bem IEEE Transactions on Information Theory}, {\Bem 62\/}(4),
  2184--2202.

\bibitem[\protect\BCAY{Bertozzi, Franco, Mohler, Short,\ \BBA\ Sledge}{Bertozzi
  et~al.}{2020}]{ref:bertozzi}
Bertozzi, A.~L., Franco, E., Mohler, G., Short, M.~B., \BBA\ Sledge, D.
  \BBOP2020\BBCP.
\newblock \BBOQ The challenges of modeling and forecasting the spread of
  covid-19\BBCQ\
\newblock {\Bem Proc Natl Acad Sci U S A}, {\Bem 117\/}(29), 16732--16738.

\bibitem[\protect\BCAY{Campbell\ \BBA\ Li}{Campbell\ \BBA\
  Li}{2019}]{ref:campbell}
Campbell, T.\BBACOMMA\  \BBA\ Li, X. \BBOP2019\BBCP.
\newblock \BBOQ Universal boosting variational inference\BBCQ\
\newblock {\Bem 33rd Conference on Neural Information Processing Systems}.

\bibitem[\protect\BCAY{Cho, Galstyan, Brantingham,\ \BBA\ Tita}{Cho
  et~al.}{2014}]{ref:cho}
Cho, Y.-S., Galstyan, A., Brantingham, P.~J., \BBA\ Tita, G. \BBOP2014\BBCP.
\newblock \BBOQ Latent self-exciting point process model for spatial-temporal
  networks\BBCQ\
\newblock {\Bem Discrete and Continuous Dynamical Systems Series B}, {\Bem
  19\/}(5), 1335--1354.

\bibitem[\protect\BCAY{Coifman\ \BBA\ Lafon}{Coifman\ \BBA\
  Lafon}{2006}]{ref:coifman}
Coifman, R.~R.\BBACOMMA\  \BBA\ Lafon, S. \BBOP2006\BBCP.
\newblock \BBOQ Diffusion maps\BBCQ\
\newblock {\Bem Applied and Computational Harmonic Analysis}, {\Bem 21}, 5--30.

\bibitem[\protect\BCAY{Cox}{Cox}{1955}]{ref:cox}
Cox, D.~R. \BBOP1955\BBCP.
\newblock \BBOQ Some statistical methods connected with series of events\BBCQ\
\newblock {\Bem Journal of the Royal Statistical Society B}, {\Bem 17\/}(2),
  129--164.

\bibitem[\protect\BCAY{Detommaso, Hoitzing, Cui,\ \BBA\ Alamir}{Detommaso
  et~al.}{2019}]{ref:detommaso}
Detommaso, G., Hoitzing, H., Cui, T., \BBA\ Alamir, A. \BBOP2019\BBCP.
\newblock \BBOQ Stein variational online changepoint detection with
  applications to hawkes processes and neural networks\BBCQ\
\newblock {\Bem arXiv:1901.07987v2}.

\bibitem[\protect\BCAY{Ding, Khan, Sato,\ \BBA\ Sugiyama}{Ding
  et~al.}{2018}]{ref:ding}
Ding, H., Khan, M.~E., Sato, I., \BBA\ Sugiyama, M. \BBOP2018\BBCP.
\newblock \BBOQ Bayesian nonparametric poisson-process allocation for
  time-sequence modeling\BBCQ\
\newblock {\Bem arXiv:1705.07006}.

\bibitem[\protect\BCAY{Drakopoulos, Ozdaglar,\ \BBA\ Tsitsiklis}{Drakopoulos
  et~al.}{2017}]{ref:drakopoulos}
Drakopoulos, K., Ozdaglar, A., \BBA\ Tsitsiklis, J.~N. \BBOP2017\BBCP.
\newblock \BBOQ When is a network epidemic hard to eliminate?\BBCQ\
\newblock {\Bem Mathematics of Operations Research}, {\Bem 42\/}(1), 1--14.

\bibitem[\protect\BCAY{Du, Dai, Trivedi, Upadhyay, Gomez-Rodriguez,\ \BBA\
  Song}{Du et~al.}{2016}]{ref:du}
Du, N., Dai, H., Trivedi, R., Upadhyay, U., Gomez-Rodriguez, M., \BBA\ Song, L.
  \BBOP2016\BBCP.
\newblock \BBOQ Recurrent marked temporal point processes: Embedding event
  history to vector\BBCQ\
\newblock {\Bem KDD}.

\bibitem[\protect\BCAY{Etesami, Kiyavash, Zhang,\ \BBA\ Singhal}{Etesami
  et~al.}{2016}]{ref:etesami}
Etesami, J., Kiyavash, N., Zhang, K., \BBA\ Singhal, K. \BBOP2016\BBCP.
\newblock \BBOQ Learning network of multivariate hawkes processes: A time
  series approach\BBCQ\
\newblock In {\Bem Proceedings of the Thirty-Second Conference on Uncertainty
  in Artificial Intelligence}.

\bibitem[\protect\BCAY{Fefferman, Mitter,\ \BBA\ Narayanan}{Fefferman
  et~al.}{2016}]{ref:fefferman}
Fefferman, C., Mitter, S., \BBA\ Narayanan, H. \BBOP2016\BBCP.
\newblock \BBOQ Testing the manifold hypothesis\BBCQ\
\newblock {\Bem J. Amer. Math. Soc.}, {\Bem 29\/}(4), 983--1049.

\bibitem[\protect\BCAY{Flaxman, Chirico, Pereira,\ \BBA\ Loeffler}{Flaxman
  et~al.}{2019}]{ref:flaxman}
Flaxman, S., Chirico, M., Pereira, P., \BBA\ Loeffler, C. \BBOP2019\BBCP.
\newblock \BBOQ Scalable high-resolution forecasting of sparse spatiotemopral
  events with kernel methods\BBCQ\
\newblock {\Bem arXiv:1801.02858}.

\bibitem[\protect\BCAY{Garske, Cori, Ariyarajah, Blake, Dorigatti, Eckmanns,
  Fraser, Hinsley, Jombart, Mills, Nedjati-Gilani, Newton, Nouvellet, Perkins,
  Riley, Schumacher, Shah, Kerkhove, Dye, Ferguson,\ \BBA\ Donnelly}{Garske
  et~al.}{2017}]{ref:tini}
Garske, T., Cori, A., Ariyarajah, A., Blake, I.~M., Dorigatti, I., Eckmanns,
  T., Fraser, C., Hinsley, W., Jombart, T., Mills, H.~L., Nedjati-Gilani, G.,
  Newton, E., Nouvellet, P., Perkins, D., Riley, S., Schumacher, D., Shah, A.,
  Kerkhove, M. D.~V., Dye, C., Ferguson, N.~M., \BBA\ Donnelly, C.~A.
  \BBOP2017\BBCP.
\newblock \BBOQ Heterogeneities in the case fatality ratio in the west african
  ebola outbreak 2013-2016\BBCQ\
\newblock {\Bem Philosophical Transactions of the Royal Society B: Biological
  Sciences}, {\Bem 372\/}(1721).

\bibitem[\protect\BCAY{Gibson, Streftaris,\ \BBA\ Thong}{Gibson
  et~al.}{2018}]{ref:gibson}
Gibson, G.~J., Streftaris, G., \BBA\ Thong, D. \BBOP2018\BBCP.
\newblock \BBOQ Comparison and assessment of epidemic models\BBCQ\
\newblock {\Bem Statistical Science}, {\Bem 33\/}(1), 19--33.

\bibitem[\protect\BCAY{Goodhart}{Goodhart}{1981}]{ref:goodhart}
Goodhart, C. \BBOP1981\BBCP.
\newblock {\Bem Inflation, Depression, and Economic Policy in the West}, \BCH\
  Problems of Monetary Management: The U.K. Experience, \BPGS\ 111--146.
\newblock Rowman \& Littlefield.

\bibitem[\protect\BCAY{Halpin\ \BBA\ Boeck}{Halpin\ \BBA\
  Boeck}{2013}]{ref:halpin}
Halpin, P.~F.\BBACOMMA\  \BBA\ Boeck, P.~D. \BBOP2013\BBCP.
\newblock \BBOQ Modelling dyadic interaction with hawkes processes\BBCQ\
\newblock {\Bem Psychometrika}, {\Bem 78\/}(4), 793--814.

\bibitem[\protect\BCAY{Hawkes}{Hawkes}{1971}]{ref:hawkes}
Hawkes, A.~G. \BBOP1971\BBCP.
\newblock \BBOQ Spectra of some self-exciting and mutually exciting point
  processes\BBCQ\
\newblock {\Bem Biometrika}, {\Bem 58\/}(1), 83--90.

\bibitem[\protect\BCAY{He, Rekatsinas, Foulds, Getoor,\ \BBA\ Liu}{He
  et~al.}{2015}]{ref:he}
He, X., Rekatsinas, T., Foulds, J., Getoor, L., \BBA\ Liu, Y. \BBOP2015\BBCP.
\newblock \BBOQ Hawkestopic: A joint model for network inference and topic
  modeling from text-based cascades\BBCQ\
\newblock {\Bem International Conference on Machine Learning}, {\Bem 32\/}(37),
  871--880.

\bibitem[\protect\BCAY{Hunter\ \BBA\ Lange}{Hunter\ \BBA\
  Lange}{2004}]{ref:lange}
Hunter, D.~R.\BBACOMMA\  \BBA\ Lange, K. \BBOP2004\BBCP.
\newblock \BBOQ A tutorial on mm algorithms\BBCQ\
\newblock {\Bem The American Statistician}, {\Bem 58\/}(1), 30--37.

\bibitem[\protect\BCAY{Jia\ \BBA\ Benson}{Jia\ \BBA\ Benson}{2019}]{ref:jia}
Jia, J.\BBACOMMA\  \BBA\ Benson, A.~R. \BBOP2019\BBCP.
\newblock \BBOQ Neural jump stochastic differential equations\BBCQ\
\newblock {\Bem 33rd Conference on Neural Information Processing Systems}.

\bibitem[\protect\BCAY{Junuthula, Haghdan, Xu,\ \BBA\ Devabhaktuni}{Junuthula
  et~al.}{2019}]{ref:junuthula}
Junuthula, R.~R., Haghdan, M., Xu, K.~S., \BBA\ Devabhaktuni, V.~K.
  \BBOP2019\BBCP.
\newblock \BBOQ The block point process model for continuous-time event-based
  dynamic networks\BBCQ\
\newblock {\Bem Proceedings of the World Wide Web Conference}.

\bibitem[\protect\BCAY{Kalatzis, Eklund, Arvanitidis,\ \BBA\ Hauberg}{Kalatzis
  et~al.}{2020}]{ref:kalatzis}
Kalatzis, D., Eklund, D., Arvanitidis, G., \BBA\ Hauberg, S. \BBOP2020\BBCP.
\newblock \BBOQ Variational autoencoders with riamannian brownian motion
  priors\BBCQ\
\newblock {\Bem arXiv:2002.05227}.

\bibitem[\protect\BCAY{Kerkhove, Bento, Mills, Ferguson,\ \BBA\
  Donnelly}{Kerkhove et~al.}{2015}]{ref:kerkhove}
Kerkhove, M. D.~V., Bento, A.~I., Mills, H.~L., Ferguson, N.~M., \BBA\
  Donnelly, C.~A. \BBOP2015\BBCP.
\newblock \BBOQ A review of epidemiological parameters from ebola outbreaks to
  inform early public health decision-making\BBCQ\
\newblock {\Bem Scientific Data}, {\Bem 2\/}(150019).

\bibitem[\protect\BCAY{Kim}{Kim}{2020}]{ref:korea}
Kim, J. \BBOP2020\BBCP.
\newblock \BBOQ Ds4c: Data science for covid-19 in south korea\BBCQ\
\newblock \BTR, Korea Centers for Disease Control \& Prevention.

\bibitem[\protect\BCAY{Lemonnier, Scaman,\ \BBA\ Kalogeratos}{Lemonnier
  et~al.}{2017}]{ref:lemonnier}
Lemonnier, R., Scaman, K., \BBA\ Kalogeratos, A. \BBOP2017\BBCP.
\newblock \BBOQ Multivariate hawkes processes for large-scale inference\BBCQ\
\newblock {\Bem Proceedings of the AAAI Conference on Artificial Intelligence}.

\bibitem[\protect\BCAY{Li, Lindenbaum, Cheng,\ \BBA\ Cloninger}{Li
  et~al.}{2019}]{ref:li}
Li, H., Lindenbaum, O., Cheng, X., \BBA\ Cloninger, A. \BBOP2019\BBCP.
\newblock \BBOQ Diffusion variational autoencoders\BBCQ\
\newblock {\Bem arXiv:1905.12724}.

\bibitem[\protect\BCAY{Lian, Talmon, Zaveri, Carin,\ \BBA\ Coifman}{Lian
  et~al.}{2015}]{ref:lian}
Lian, W., Talmon, R., Zaveri, H., Carin, L., \BBA\ Coifman, R. \BBOP2015\BBCP.
\newblock \BBOQ Multivariate time-series analysis and diffusion maps\BBCQ\
\newblock {\Bem Signal Processing}, {\Bem 116}, 13--28.

\bibitem[\protect\BCAY{Linderman\ \BBA\ Adams}{Linderman\ \BBA\
  Adams}{2015}]{ref:linderman}
Linderman, S.~W.\BBACOMMA\  \BBA\ Adams, R.~P. \BBOP2015\BBCP.
\newblock \BBOQ Scalable bayesian inference for excitatory point process
  networks\BBCQ\
\newblock {\Bem arXiv:1507.03228}.

\bibitem[\protect\BCAY{Liu\ \BBA\ Hauskrecht}{Liu\ \BBA\
  Hauskrecht}{2019}]{ref:liu}
Liu, S.\BBACOMMA\  \BBA\ Hauskrecht, M. \BBOP2019\BBCP.
\newblock \BBOQ Nonparametric regressive point processes based on conditional
  gaussian processes\BBCQ\
\newblock {\Bem 33rd Conference on Neural Information Processing Systems}.

\bibitem[\protect\BCAY{Lloyd, Gunter, Osborne, Roberts,\ \BBA\ Nickson}{Lloyd
  et~al.}{2016}]{ref:lloyd}
Lloyd, C., Gunter, T., Osborne, M., Roberts, S., \BBA\ Nickson, T.
  \BBOP2016\BBCP.
\newblock \BBOQ Latent point process allocation\BBCQ\
\newblock {\Bem International Conference on Artificial Intelligence and
  Statistics}, {\Bem 19\/}(51), 389--397.

\bibitem[\protect\BCAY{Loaiza-Ganem, Perkins, Schroeder, Churchland,\ \BBA\
  Cunningham}{Loaiza-Ganem et~al.}{2019}]{ref:ganem}
Loaiza-Ganem, G., Perkins, S.~M., Schroeder, K.~E., Churchland, M.~M., \BBA\
  Cunningham, J.~P. \BBOP2019\BBCP.
\newblock \BBOQ Deep random splines for point process intensity estimation of
  neural population data\BBCQ\
\newblock {\Bem 33rd Conference on Neural Information Processing Systems}.

\bibitem[\protect\BCAY{Lu, Wang, Shi, Yu,\ \BBA\ Ye}{Lu et~al.}{2019}]{ref:lu}
Lu, Y., Wang, X., Shi, C., Yu, P.~S., \BBA\ Ye, Y. \BBOP2019\BBCP.
\newblock \BBOQ Temporal network embedding with micro- and macro-dynamics\BBCQ\
\newblock {\Bem Proceedings of the Conference of Information and Knowledge
  Management}.

\bibitem[\protect\BCAY{Marmarelis\ \BBA\ Berger}{Marmarelis\ \BBA\
  Berger}{2005}]{ref:marmarelis}
Marmarelis, V.~Z.\BBACOMMA\  \BBA\ Berger, T.~W. \BBOP2005\BBCP.
\newblock \BBOQ General methodology for nonlinear modeling of neural systems
  with poisson point-process inputs\BBCQ\
\newblock {\Bem Mathematical Biosciences}, {\Bem 196}, 1--13.

\bibitem[\protect\BCAY{McFadden}{McFadden}{1965}]{ref:mcfadden}
McFadden, J.~A. \BBOP1965\BBCP.
\newblock \BBOQ The entropy of a point process\BBCQ\
\newblock {\Bem Society for Industrial and Applied Mathematics}, {\Bem
  13\/}(4), 988--994.

\bibitem[\protect\BCAY{Mehrasa, Deng, Ahmed, Chang, He, Durand, Brubaker,\
  \BBA\ Mori}{Mehrasa et~al.}{2019}]{ref:mehrasa}
Mehrasa, N., Deng, R., Ahmed, M.~O., Chang, B., He, J., Durand, T., Brubaker,
  M., \BBA\ Mori, G. \BBOP2019\BBCP.
\newblock \BBOQ Point process flows\BBCQ\
\newblock {\Bem arXiv:1910.08281}.

\bibitem[\protect\BCAY{Mei\ \BBA\ Eisner}{Mei\ \BBA\ Eisner}{2017}]{ref:mei}
Mei, H.\BBACOMMA\  \BBA\ Eisner, J. \BBOP2017\BBCP.
\newblock \BBOQ The neural hawkes process: A neurally self-modulating
  multivatiate point process\BBCQ\
\newblock {\Bem 31st Conference on Neural Information Processing Systems}.

\bibitem[\protect\BCAY{Mohler, Short, Brantingham, Schoenberg,\ \BBA\
  Tita}{Mohler et~al.}{2011}]{ref:mohler2011}
Mohler, G.~O., Short, M.~B., Brantingham, P.~J., Schoenberg, F.~P., \BBA\ Tita,
  G.~E. \BBOP2011\BBCP.
\newblock \BBOQ Self-exciting point process modeling of crime\BBCQ\
\newblock {\Bem Journal of the American Statistical Association}, {\Bem
  106\/}(493), 100--108.

\bibitem[\protect\BCAY{Mohler}{Mohler}{2014}]{ref:mohler}
Mohler, G. \BBOP2014\BBCP.
\newblock \BBOQ Marked point process hotspot maps for homicide and gun crime
  prediction in chicago\BBCQ\
\newblock {\Bem International Journal of Forecasting}, {\Bem 30}, 491--497.

\bibitem[\protect\BCAY{Morariu-Patrichi\ \BBA\ Pakkanen}{Morariu-Patrichi\
  \BBA\ Pakkanen}{2018}]{ref:pakkanen}
Morariu-Patrichi, M.\BBACOMMA\  \BBA\ Pakkanen, M.~S. \BBOP2018\BBCP.
\newblock \BBOQ State-dependent hawkes processes and their application to limit
  order book modeling\BBCQ\
\newblock {\Bem arXiv:1809.08060v2}.

\bibitem[\protect\BCAY{Nesterov\ \BBA\ Polyak}{Nesterov\ \BBA\
  Polyak}{2006}]{ref:nesterov}
Nesterov, Y.\BBACOMMA\  \BBA\ Polyak, B. \BBOP2006\BBCP.
\newblock \BBOQ Cubic regularization of newton method and its global
  performance\BBCQ\
\newblock {\Bem Math. Program.}, {\Bem A\/}(108), 177--205.

\bibitem[\protect\BCAY{Newson}{Newson}{2002}]{ref:newson}
Newson, R. \BBOP2002\BBCP.
\newblock \BBOQ Parameters behind ``nonparametric'' statistics: Kendall's tau,
  somers' d and median differences\BBCQ\
\newblock {\Bem The State Journal}, {\Bem 2\/}(1), 45--64.

\bibitem[\protect\BCAY{Nickel\ \BBA\ Le}{Nickel\ \BBA\ Le}{2020}]{ref:nickel}
Nickel, M.\BBACOMMA\  \BBA\ Le, M. \BBOP2020\BBCP.
\newblock \BBOQ Learning multivariate hawkes processes at scale\BBCQ\
\newblock {\Bem arXiv:2002.12501}.

\bibitem[\protect\BCAY{Ogata}{Ogata}{1981}]{ref:ogata}
Ogata, Y. \BBOP1981\BBCP.
\newblock \BBOQ On lewis’ simulation method for point processes\BBCQ\
\newblock {\Bem IEEE Transactions on Information Theory}, {\Bem IT-27\/}(1),
  23--31.

\bibitem[\protect\BCAY{Okawa, Iwata, Kurashima, Tanaka, Toda,\ \BBA\
  Ueda}{Okawa et~al.}{2019}]{ref:okawa}
Okawa, M., Iwata, T., Kurashima, T., Tanaka, Y., Toda, H., \BBA\ Ueda, N.
  \BBOP2019\BBCP.
\newblock \BBOQ Deep mixture point processes: Spatio-temporal event prediction
  with rich contextual information\BBCQ\
\newblock {\Bem KDD}.

\bibitem[\protect\BCAY{Oliver, Lepri, Sterly, Lambiotte, Delataille, De~Nadai,
  Letouz{\'e}, Salah, Benjamins, Cattuto, Colizza, de~Cordes, Fraiberger,
  Koebe, Lehmann, Murillo, Pentland, Pham, Pivetta, Saram{\"a}ki, Scarpino,
  Tizzoni, Verhulst,\ \BBA\ Vinck}{Oliver et~al.}{2020}]{ref:oliver}
Oliver, N., Lepri, B., Sterly, H., Lambiotte, R., Delataille, S., De~Nadai, M.,
  Letouz{\'e}, E., Salah, A.~A., Benjamins, R., Cattuto, C., Colizza, V.,
  de~Cordes, N., Fraiberger, S.~P., Koebe, T., Lehmann, S., Murillo, J.,
  Pentland, A., Pham, P.~N., Pivetta, F., Saram{\"a}ki, J., Scarpino, S.~V.,
  Tizzoni, M., Verhulst, t., \BBA\ Vinck, P. \BBOP2020\BBCP.
\newblock \BBOQ Mobile phone data for informing public health actions across
  the covid-19 pandemic life cycle\BBCQ\
\newblock {\Bem Science Advances}.

\bibitem[\protect\BCAY{Omi}{Omi}{2019}]{ref:omi}
Omi, T. \BBOP2019\BBCP.
\newblock \BBOQ Fully neural network based model for general temporal point
  processes\BBCQ\
\newblock {\Bem 33rd Conference on Neural Information Processing Systems}.

\bibitem[\protect\BCAY{Pennington, Socher,\ \BBA\ Manning}{Pennington
  et~al.}{2014}]{ref:pennington}
Pennington, J., Socher, R., \BBA\ Manning, C.~D. \BBOP2014\BBCP.
\newblock \BBOQ Glove: Global vectors for word representation\BBCQ\
\newblock {\Bem Conference on Empirical Methods in Natural Language
  Processing}.

\bibitem[\protect\BCAY{Pham\ \BBA\ Chen}{Pham\ \BBA\ Chen}{2018}]{ref:pham}
Pham, K.\BBACOMMA\  \BBA\ Chen, G. \BBOP2018\BBCP.
\newblock \BBOQ Large-scale spectral clustering using diffusion coordinates on
  landmark-based bipartite graphs\BBCQ\
\newblock {\Bem Workshop on Graph-Based Methods for Natural Language
  Processing}, {\Bem 12}, 28--37.

\bibitem[\protect\BCAY{Pillow, Shlens, Paninski, Sher, Litke, Chichilnisky,\
  \BBA\ Simoncelli}{Pillow et~al.}{2008}]{ref:pillow}
Pillow, J.~W., Shlens, J., Paninski, L., Sher, A., Litke, A.~M., Chichilnisky,
  E.~J., \BBA\ Simoncelli, E.~P. \BBOP2008\BBCP.
\newblock \BBOQ Spatio-temporal correlations and visual signaling in a complete
  neuronal population\BBCQ\
\newblock {\Bem Nature}, {\Bem 454}, 995--999.

\bibitem[\protect\BCAY{Razali\ \BBA\ Wah}{Razali\ \BBA\ Wah}{2011}]{ref:razali}
Razali, N.~M.\BBACOMMA\  \BBA\ Wah, Y.~B. \BBOP2011\BBCP.
\newblock \BBOQ Power comparisons of shapiro-wilk, kolmogorov-smirnov,
  lilliefors and anderson-darling tests\BBCQ\
\newblock {\Bem Journal of Statistical Modeling and Analytics}, {\Bem 2\/}(1),
  21--33.

\bibitem[\protect\BCAY{Reinhart}{Reinhart}{2018}]{ref:reinhart}
Reinhart, A. \BBOP2018\BBCP.
\newblock \BBOQ A review of self-exciting spatio-temporal point processes and
  their applications\BBCQ\
\newblock {\Bem Statistical Science}, {\Bem 33\/}(3), 299--318.

\bibitem[\protect\BCAY{Rey, Menkovski,\ \BBA\ Portegies}{Rey
  et~al.}{2019}]{ref:rey}
Rey, L. A.~P., Menkovski, V., \BBA\ Portegies, J.~W. \BBOP2019\BBCP.
\newblock \BBOQ Diffusion variational autoencoders\BBCQ\
\newblock {\Bem arXiv:1901.08991}.

\bibitem[\protect\BCAY{Salehi, Trouleau, Grossglauser,\ \BBA\ Thiran}{Salehi
  et~al.}{2019}]{ref:salehi}
Salehi, F., Trouleau, W., Grossglauser, M., \BBA\ Thiran, P. \BBOP2019\BBCP.
\newblock \BBOQ Learning hawkes processes from a handful of events\BBCQ\
\newblock {\Bem 33rd Conference on Neural Information Processing Systems}.

\bibitem[\protect\BCAY{Shang\ \BBA\ Sun}{Shang\ \BBA\ Sun}{2019}]{ref:shang}
Shang, J.\BBACOMMA\  \BBA\ Sun, M. \BBOP2019\BBCP.
\newblock \BBOQ Geometric hawkes processes with graph convolutional recurrent
  neural networks\BBCQ\
\newblock {\Bem Association for the Advancement of Artificial Intelligence}.

\bibitem[\protect\BCAY{Sharma, Ghosh,\ \BBA\ Fiterau}{Sharma
  et~al.}{2019}]{ref:sharma}
Sharma, A., Ghosh, A., \BBA\ Fiterau, M. \BBOP2019\BBCP.
\newblock \BBOQ Generative sequential stochastic model for marked point
  processes\BBCQ\
\newblock {\Bem ICML Time Series Workshop}.

\bibitem[\protect\BCAY{Soize\ \BBA\ Ghanem}{Soize\ \BBA\
  Ghanem}{2016}]{ref:ghanem}
Soize, C.\BBACOMMA\  \BBA\ Ghanem, R. \BBOP2016\BBCP.
\newblock \BBOQ Data-driven probability concentration and sampling on
  manifold\BBCQ\
\newblock {\Bem Journal of Computational Physics}, {\Bem 321}, 242--258.

\bibitem[\protect\BCAY{Student}{Student}{1908}]{ref:student}
Student \BBOP1908\BBCP.
\newblock \BBOQ The probable error of a mean\BBCQ\
\newblock {\Bem Biometrika}, {\Bem 6\/}(1).

\bibitem[\protect\BCAY{Swishchuk\ \BBA\ Huffman}{Swishchuk\ \BBA\
  Huffman}{2020}]{ref:swishchuk}
Swishchuk, A.\BBACOMMA\  \BBA\ Huffman, A. \BBOP2020\BBCP.
\newblock \BBOQ General compound hawkes processes in limit order books\BBCQ\
\newblock {\Bem Risks}, {\Bem 8\/}(28).

\bibitem[\protect\BCAY{Taylor, Klimm, Harrington, Kramár, Mischaikow, Porter,\
  \BBA\ Mucha}{Taylor et~al.}{2015}]{ref:taylor}
Taylor, D., Klimm, F., Harrington, H.~A., Kramár, M., Mischaikow, K., Porter,
  M.~A., \BBA\ Mucha, P.~J. \BBOP2015\BBCP.
\newblock \BBOQ Topological data analysis of contagion maps for examining
  spreading processes on networks\BBCQ\
\newblock {\Bem Nat Commun}, {\Bem 6}, 7723.

\bibitem[\protect\BCAY{Torricelli, Karsai,\ \BBA\ Gauvin}{Torricelli
  et~al.}{2020}]{ref:torricelli}
Torricelli, M., Karsai, M., \BBA\ Gauvin, L. \BBOP2020\BBCP.
\newblock \BBOQ weg2vec: Event embedding for temporal networks\BBCQ\
\newblock {\Bem Scientific Reports}, {\Bem 10\/}(7164).

\bibitem[\protect\BCAY{Veen\ \BBA\ Schoenberg}{Veen\ \BBA\
  Schoenberg}{2008}]{ref:schoenberg}
Veen, A.\BBACOMMA\  \BBA\ Schoenberg, F.~P. \BBOP2008\BBCP.
\newblock \BBOQ Estimation of space-time branching process models in seismology
  using an em-type algorithm\BBCQ\
\newblock {\Bem Journal of the American Statistical Association}, {\Bem
  103\/}(482), 614--624.

\bibitem[\protect\BCAY{Welsh}{Welsh}{2020}]{ref:latimes}
Welsh, B. \BBOP2020\BBCP.
\newblock \BBOQ Los angeles times\BBCQ\
\newblock \url{https://github.com/datadesk/california-coronavirus-data}.

\bibitem[\protect\BCAY{Xu, Farajtabat,\ \BBA\ Zha}{Xu et~al.}{2016}]{ref:xu}
Xu, H., Farajtabat, M., \BBA\ Zha, H. \BBOP2016\BBCP.
\newblock \BBOQ Learning granger causality for hawkes processes\BBCQ\
\newblock {\Bem Proceedings of the 33rd International Conference on Machine
  Learning}, {\Bem 48}.

\bibitem[\protect\BCAY{Yang, Kutan,\ \BBA\ Ryu}{Yang et~al.}{2018}]{ref:yang}
Yang, H., Kutan, A.~M., \BBA\ Ryu, D. \BBOP2018\BBCP.
\newblock \BBOQ Option moneyness and price disagreements\BBCQ\
\newblock {\Bem Applied Economics Letters}, {\Bem 25\/}(3), 192--196.

\bibitem[\protect\BCAY{Yuan, Li, Bertozzi, Brantingham,\ \BBA\ Porter}{Yuan
  et~al.}{2019}]{ref:yuan}
Yuan, B., Li, H., Bertozzi, A.~L., Brantingham, P.~J., \BBA\ Porter, M.~A.
  \BBOP2019\BBCP.
\newblock \BBOQ Multivariate spatiotemporal hawkes processes and network
  reconstruction\BBCQ\
\newblock {\Bem SIAM J. Math. Data Sci.}, {\Bem 1\/}(2), 356--382.

\bibitem[\protect\BCAY{Zhang, Walder, Rizoiu,\ \BBA\ Xie}{Zhang
  et~al.}{2019}]{ref:zhang}
Zhang, R., Walder, C., Rizoiu, M.-A., \BBA\ Xie, L. \BBOP2019\BBCP.
\newblock \BBOQ Efficient non-parametric bayesian hawkes processes\BBCQ\
\newblock {\Bem International Joint Conference on Artificial Intelligence},
  {\Bem 28}.

\bibitem[\protect\BCAY{Zhou, Zha,\ \BBA\ Song}{Zhou et~al.}{2013a}]{ref:zhou1}
Zhou, K., Zha, H., \BBA\ Song, L. \BBOP2013a\BBCP.
\newblock \BBOQ Learning social infectivity in sparse low-rank networks using
  multi-dimensional hawkes processes\BBCQ\
\newblock {\Bem 16th International Conference on Artificial Intelligence and
  Statistics}, {\Bem 31}.

\bibitem[\protect\BCAY{Zhou, Zha,\ \BBA\ Song}{Zhou et~al.}{2013b}]{ref:zhou2}
Zhou, K., Zha, H., \BBA\ Song, L. \BBOP2013b\BBCP.
\newblock \BBOQ Learning triggering kernels for multi-dimensional hawkes
  processes\BBCQ\
\newblock {\Bem 30th International Conference on Machine Learning}.

\bibitem[\protect\BCAY{Zhu, Li,\ \BBA\ Xie}{Zhu et~al.}{2020}]{ref:zhu}
Zhu, S., Li, S., \BBA\ Xie, Y. \BBOP2020\BBCP.
\newblock \BBOQ Interpretable generative neural spatio-temporal point
  processes\BBCQ\
\newblock {\Bem arXiv:1906.05467v2}.

\bibitem[\protect\BCAY{Zhu\ \BBA\ Xie}{Zhu\ \BBA\ Xie}{2019}]{ref:zhu2}
Zhu, S.\BBACOMMA\  \BBA\ Xie, Y. \BBOP2019\BBCP.
\newblock \BBOQ Crime event embedding with unsupervised feature selection\BBCQ\
\newblock {\Bem Proceedings of the International Conference on Acoustics,
  Speech, and Signal Processing}.

\bibitem[\protect\BCAY{Zuo, Liu, Lin, Guo, Hu,\ \BBA\ Wu}{Zuo
  et~al.}{2018}]{ref:zuo}
Zuo, Y., Liu, G., Lin, H., Guo, J., Hu, X., \BBA\ Wu, J. \BBOP2018\BBCP.
\newblock \BBOQ Embedding temporal network via neighborhood formation\BBCQ\
\newblock {\Bem KDD}.

\end{thebibliography}

\end{document}